\documentclass{article}
\usepackage{venues/iclr/configuration/iclr_conference}
\iclrfinalcopy
\usepackage{times}
\usepackage[T1]{fontenc}
\usepackage[utf8]{inputenc}
\usepackage{url}
\usepackage{graphicx}
\usepackage{booktabs}
\usepackage{multirow}
\usepackage{tabularx}
\usepackage{placeins}
\usepackage{dblfloatfix}
\usepackage{float}
\usepackage{comment}
\usepackage{enumitem}
\usepackage{xcolor}
\usepackage{colortbl}
\usepackage{hyperref}
\usepackage[most]{tcolorbox}
\usepackage{pifont}
\usepackage{microtype}
\usepackage{inconsolata}
\usepackage{xspace}
\usepackage[normalem]{ulem}
\usepackage{common/support/algorithm}
\usepackage{common/support/algorithmicx}
\usepackage[noend]{common/support/algpseudocode}
\usepackage{amsmath,amsthm,amssymb}
\newtheorem{theorem}{Theorem}[section]

\newtheorem{definition}[theorem]{Definition}

\newenvironment{talign*}
{\csname align*\endcsname}
{\endalign}
\usepackage{algorithm}
\usepackage[noend]{algpseudocode}

\algnewcommand\algorithmicforeach{\textbf{for each}}
\algdef{S}[FOR]{ForEach}[1]{\algorithmicforeach\ #1\ \algorithmicdo}

\errorcontextlines\maxdimen

\makeatletter
\newcommand*{\algrule}[1][\algorithmicindent]{\makebox[#1][l]{\hspace*{.5em}\thealgruleextra\vrule height \thealgruleheight depth \thealgruledepth}}%
\newcommand*{\thealgruleextra}{}
\newcommand*{\thealgruleheight}{.75\baselineskip}
\newcommand*{\thealgruledepth}{.25\baselineskip}

\newcount\ALG@printindent@tempcnta
\def\ALG@printindent{%
	\ifnum \theALG@nested>0
	\ifx\ALG@text\ALG@x@notext
	\else
		\unskip
		\addvspace{-1pt}
		\ALG@printindent@tempcnta=1
		\loop
		\algrule[\csname ALG@ind@\the\ALG@printindent@tempcnta\endcsname]%
		\advance \ALG@printindent@tempcnta 1
		\ifnum \ALG@printindent@tempcnta<\numexpr\theALG@nested+1\relax
			\repeat
		\fi
	\fi
}%
\usepackage{etoolbox}
\patchcmd{\ALG@doentity}{\noindent\hskip\ALG@tlm}{\ALG@printindent}{}{\errmessage{failed to patch}}
\makeatother

\newbox\statebox
\newcommand{\myState}[1]{%
	\setbox\statebox=\vbox{#1}%
	\edef\thealgruleheight{\dimexpr \the\ht\statebox+1pt\relax}%
	\edef\thealgruledepth{\dimexpr \the\dp\statebox+1pt\relax}%
	\ifdim\thealgruleheight<.75\baselineskip
		\def\thealgruleheight{\dimexpr .75\baselineskip+1pt\relax}%
	\fi
	\ifdim\thealgruledepth<.25\baselineskip
		\def\thealgruledepth{\dimexpr .25\baselineskip+1pt\relax}%
	\fi
	\State #1%
	\def\thealgruleheight{\dimexpr .75\baselineskip+1pt\relax}%
	\def\thealgruledepth{\dimexpr .25\baselineskip+1pt\relax}%
}

\graphicspath{{common/}{common/figures/}{./}}

\theoremstyle{definition}

\newcommand{\filled}[1]{\uline{#1}}
\definecolor{darkgray}{gray}{0.25}
\definecolor{taggreen}{RGB}{0,128,0}
\definecolor{tagmagenta}{RGB}{170,0,120}
\definecolor{tagteal}{RGB}{0,110,130}

\newcommand{\thinktag}[1]{\textcolor{taggreen}{\texttt{\textless #1\textgreater}}}
\newcommand{\actiontag}[1]{\textcolor{tagmagenta}{\texttt{\textless #1\textgreater}}}

\newenvironment{promptbox}[1]{%
  \begin{tcolorbox}[
    enhanced,
    colback=white,
    colframe=black,
    colbacktitle=black,
    coltitle=white,
    title={#1},
    fonttitle=\bfseries,
    boxrule=0.8pt,
    arc=6pt,
    left=10pt,
    right=10pt,
    top=8pt,
    bottom=8pt,
    titlerule=0pt,
    before skip=6pt,
    after skip=6pt
  ]
}{%
  \end{tcolorbox}
}

\newcommand{\PaperTitle}{ProAS: A Progress-Aware Advantage Shaping Framework for Multi-Turn Agentic Reinforcement Learning}

\newcommand{\PaperAuthors}{Bowen Ding, Mingda Wan, Yuhan Chen, Jiyao Yuan, Qi Zhu, Futing Wang, Heyuan Deng, Fei Mi, Lifeng Shang, Tao Lin}

\makeatletter
\newcommand{\PaperIclrAuthorBlock}{
    Bowen Ding\textsuperscript{\rm 1,2}~,
    Mingda Wan\textsuperscript{\rm 2}~,
    Yuhan Chen\textsuperscript{\rm 2,*}~,
    Jiyao Yuan\textsuperscript{\rm 4,\dag}~,
    Qi Zhu\textsuperscript{\rm 4,\dag}~,
    Futing Wang\textsuperscript{\rm 1,2}~,
  \\
  \bf Heyuan Deng\textsuperscript{\rm 4}~,
    Fei Mi\textsuperscript{\rm 4,\ddag}~,
    Lifeng Shang\textsuperscript{\rm 4}~,
    Tao Lin\textsuperscript{\rm 2,3,\ddag}~,
  \\ [1mm]
    \textsuperscript{1} Zhejiang University~
    \textsuperscript{2} School of Engineering, Westlake University \\
    \textsuperscript{3} Institute of Advanced Technology, Westlake Institute for Advanced Study \\
    \textsuperscript{4} Huawei Noah's Ark Lab \\
    $^{2}$\texttt{\{dingbowen, wanmingda, wangfuting, lintao\}@westlake.edu.cn} \\
    $^{4}$\texttt{\{dengheyuan, mifei2, Shang.Lifeng\}@huawei.com}%
    \gdef\@thanks{%
      \begingroup
      \long\def\@makefntext####1{\noindent ####1}%
      \footnotetext{%
        $^{*}$Work done at Westlake as an intern.\quad
        $^{\dagger}$Work done at Huawei Noah's Ark Lab.\quad
        $^{\ddagger}$Corresponding authors.}%
      \endgroup
    }%
}
\makeatother

\begin{document}
\title{\PaperTitle}
\author{\PaperIclrAuthorBlock}
\hypersetup{
  pdftitle={\PaperTitle},
  pdfauthor={\PaperAuthors},
  pdfsubject={Preprint}
}
\maketitle
\lhead{Preprint}

\begin{abstract}
Reinforcement learning (RL) has emerged as the primary paradigm for optimizing Large Language Model (LLM) agents, yet efficient critic-free algorithms (e.g., GRPO) suffer from a fundamental \textbf{\textit{uniform credit assignment dilemma}} that broadcasts trajectory-level advantages equally to all actions. Our analysis reveals that this uniformity inherently reinforces structurally unreasonable actions in preferred trajectories and ignores each action's downstream deviations in unpreferred ones. To resolve this without introducing critic overhead, we propose \textbf{Progress-Aware Advantage Shaping (ProAS)}, a plug-and-play framework that leverages a dynamically evolving Agentic Reference Graph (ARG) to achieve fine-grained credit assignment via Masking Unreasonable Incentives (MUI) and Asymmetric Penalty Redistribution (APR). Extensive experiments across Qwen2.5-1.5B-Instruct and Qwen2.5-7B-Instruct demonstrate that ProAS universally elevates mainstream group-based baselines to new state-of-the-art performance, achieving peak success rates of 91.9\% and 94.3\% on ALFWorld, alongside 66.7\% and 70.8\% on WebShop, respectively. Furthermore, it accelerates training convergence by up to $2.44\times$ while maintaining strong robustness against imperfect or distilled ARGs. Codes: \url{https://github.com/LINs-lab/ProAS}.
\end{abstract}

\section{Introduction} \label{sec:introduction}

Large language models (LLMs) are increasingly deployed as autonomous agents that make sequential decisions through interactions with external environments. Their planning and reasoning capabilities have enabled applications in software engineering~\citep{swe-agent}, personal assistance~\citep{assistantbench}, and graphical user interface navigation~\citep{luo2025gui}. 

Reinforcement Learning (RL) has emerged as the primary paradigm for elevating the agentic capacity of LLMs~\citep{glm5,deepseek-v4}.
During RL rollouts, agents generate trajectories composed of interleaved environment states and actions, which introduces a fundamental \textbf{credit assignment challenge}:
\begin{tcolorbox}[
    colback=red!5!white,
    colframe=red!45!black,
    boxrule=0.6pt,
    arc=2mm,
    left=1mm,
    right=1mm,
    top=0.8mm,
    bottom=0.8mm,
    before skip=5pt,
    after skip=5pt
]
    \centering
    \emph{How should trajectory-level rewards and penalties be attributed to individual actions?}
\end{tcolorbox}

Actor-critic algorithms (e.g., PPO~\citep{ppo}) address this by learning a separate critic, which introduces substantial computational overhead and optimization instability in long-horizon settings~\citep{sppo,vapo}. Conversely, critic-free algorithms like Group Relative Policy Optimization (GRPO)~\citep{shao2024deepseekmathpushinglimitsmathematical} offer computational efficiency by broadcasting a trajectory-level advantage uniformly to every action~\citep{ding2025thinkingtokenshelptrap}. However, this inherently treats all intra-trajectory actions identically, completely overlooking their critical differences in local validity, task contribution, and downstream influence. As detailed in \S~\ref{sec:grpo}, we term this structural limitation the \textbf{\textit{uniform credit assignment dilemma}}.

To systematically understand the uniform credit assignment dilemma, we conduct a fine-grained action-intervention analysis (\S~\ref{sec:action_intervention_analysis}). Our findings reveal a dual suboptimality. First, in preferred trajectories with positive advantages, uniform credit inadvertently reinforces structurally invalid actions (i.e., unreasonable actions) that pollute the context history and impede the agent from reaching success. Second, in unpreferred trajectories with negative advantages, uniform penalization ignores the autoregressive nature of LLMs, where the later  deviations may drive from earlier actions. Consequently, they suppress all actions equally but arbitrarily without considering their actual downstream influence.

To resolve this dilemma without introducing critic overhead, we propose Progress-Aware Advantage Shaping (ProAS) in \S~\ref{sec:proas_framework}. ProAS leverages an Agentic Reference Graph (ARG, mentioned in \S~\ref{sec:arg_construction}), which abstracts task progression into dynamically evolving sub-goals to formally annotate whether an action is \emph{reasonable} (structurally valid) or \emph{progress-advancing} (driving task completion). Using these structural annotations, ProAS asymmetrically reshapes trajectory-level advantages in Eq.~\ref{eq:proas_advantage_shaping}: \textbf{(1) Masking Unreasonable Incentives (MUI)} zeros out the local advantages of \emph{unreasonable} actions in preferred trajectories to prevent their inadvertent reinforcement. \textbf{(2) Asymmetric Penalty Redistribution (APR)} utilizes Future-Drift (Eq.~\ref{eq:future_drift}), the inclusive count of downstream \emph{non-advancing} actions, to assign proportionally heavier penalties to actions triggering severe downstream stagnation.

Extensive experiments in \S~\ref{sec:experimental_results} on ALFWorld~\citep{shridhar2021alfworld} and WebShop~\citep{yao2023webshopscalablerealworldweb} using Qwen2.5-1.5B-Instruct and Qwen2.5-7B-Instruct~\citep{qwen25technicalreport} validate ProAS as a versatile, plug-and-play enhancer. It universally elevates mainstream group-based RL baselines (GRPO~\citep{shao2024deepseekmathpushinglimitsmathematical}, DAPO~\citep{yu2025dapoopensourcellmreinforcement}, GiGPO~\citep{feng2025group}), boosting success rates by up to 16.7 points (e.g., GRPO on 1.5B ALFWorld) and accelerating training convergence by $1.12\times$ to $2.44\times$. On top of that, this orthogonal enhancement propels these standard algorithms to establish new state-of-the-art (SOTA) performance, consistently eclipsing the strong segment-level baseline, MiGPO~\citep{MiGPO}. Notably, GiGPO+ProAS outperforms MiGPO by a massive 13.1 points on WebShop with 1.5B model and secures the highest overall success rate of 94.3\% on ALFWorld with 7B model. Furthermore, comprehensive ablations (\S~\ref{sec:ablation_studies}) show that ProAS's core mechanisms are both synergistic and superior to alternative advantage shaping variants, and confirm ProAS's robust efficacy against incomplete or distilled ARGs, underscoring the necessity of ARG's online evolution mechanism.

In summary, our primary contributions are:
\begin{itemize}[leftmargin=12pt]
    \item \textbf{Problem Identification:} We expose the uniform credit assignment dilemma in critic-free agentic RL, demonstrating how uniform advantages inadvertently reinforce contextual noise in preferred trajectories and ignore the downstream influence of actions in unpreferred ones.
    \item \textbf{Algorithmic Innovation:} We propose Progress-Aware Advantage Shaping (ProAS), a plug-and-play framework. Guided by a dynamically evolving Agentic Reference Graph (ARG), ProAS achieves fine-grained, critic-free credit assignment via Masking Unreasonable Incentives (MUI) and Asymmetric Penalty Redistribution (APR).
    \item \textbf{Empirical Superiority:} ProAS acts as a universal enhancer across multiple model scales and environments. It significantly accelerates training convergence, propels standard group-based algorithms to surpass strong SOTA baselines (e.g., MiGPO), and exhibits high robustness against imperfect and distilled  reference guidance.  
\end{itemize}

\section{Related Works} \label{sec:related_work}
\noindent \textbf{Multi-Turn Agentic Tasks.}
Large Language Model (LLM) agents are increasingly deployed to solve complex real-world tasks~\citep{TheAgentCompany}. Given the intricacies of these environments, agents must engage in multi-turn interactions, continuously accumulating contextual knowledge and making sequential decisions to drive task completion. To facilitate the training and evaluation of such agents, the community has introduced various simulated environments. Examples include ALFWorld~\citep{shridhar2021alfworld} for embodied household tasks, WebShop~\citep{yao2023webshopscalablerealworldweb} for web navigation, AppWorld~\citep{appworld} for application development, and $\tau^2$-Bench~\citep{barres2025tau2} for evaluating dynamic conversational abilities in domains like aviation and retail.

Rather than constructing new environments, our work focuses on developing more efficient and effective training paradigms tailored for existing multi-turn agentic tasks.

\noindent \textbf{RL for LLM Agents.}
Critic-free RL like GRPO~\citep{shao2024deepseekmathpushinglimitsmathematical} is widely adopted for LLM agents~\citep{toolrl2025,search-r1,wei2025webagent-r1}, but suffers from sparse-reward advantage vanishment~\citep{yu2025dapoopensourcellmreinforcement} and coarse trajectory-level credit assignment~\citep{hcapo2026}. While DAPO~\citep{yu2025dapoopensourcellmreinforcement} and dense-reward methods~\citep{toolrl2025, lu2026dontjustfinetuneagent, E-GRPO} mitigate this vanishment, their credit assignment remains trajectory-bound. To differentiate intra-trajectory actions, GiGPO~\citep{feng2025group} and MiGPO~\citep{MiGPO} compute relative advantages by matching anchor states or milestones across rollouts. 

In this work, we systematically expose the flaws of uniform credit assignment in both preferred and unpreferred trajectories, addressing them through  advantage shaping, which serves as a plug-and-play module for existing critic-free RL algorithms.


\section{Preliminary} 
\label{sec:preliminary}
\subsection{Decision Process of LLM Agents} 
\label{sec:decision_process_of_llm_agents}
Given a task context $\mathbf{x} \sim p(X)$, an LLM agent aims to solve the task within a maximum horizon of $T$ steps. At each step $t \in [1, T]$, the agent receives an immediate environment state $s_t$. Governed by a parameterized policy $\pi_{\theta}$, the agent samples an action $a_t$ from the admissible action space $\mathcal{A}$~\citep{SuttonB98}, conditioned on the task context $\mathbf{x}$, the interaction history $h_{t-1} = \{(s_i, a_i)\}_{i=1}^{t-1}$, and the current state $s_t$. Formally, this decision-making process is defined as:
\begin{equation}
    \label{eq:action_sampling}
    a_t \sim \pi_{\theta}(\cdot \mid \mathbf{x}, h_{t-1}, s_t), \quad a_t \in \mathcal{A}, \quad t \in [1, T]
\end{equation}
In practice, some harnessing wrappers~\citep{feng2025group} often provide partial observations or integrate summarized historical information~\citep{CompactionRL} (denoted as $\hat{h}_{t-1}$) into the context prior to the current state $s_t$.

\subsection{GRPO}
\label{sec:grpo}
GRPO~\citep{shao2024deepseekmathpushinglimitsmathematical} utilizes a group-level normalization strategy to estimate the advantage for assessing the action value. In GRPO, given a task $\mathbf{x}$, the old policy $\pi_{\theta_{\mathrm{old}}}$ generates a group of $G$ rollout trajectories $\{\tau_i\}_{i=1}^G$, each evaluated by a verifiable outcome reward $R(\tau_i)$. The trajectory-level advantage $A_i$ for $\tau_i$ is computed by normalizing the rewards within the group:
\begin{equation}
    \label{eq:group_relative_advantage}
    \textcolor{red}{A_i} = \frac{R(\tau_i)-\mu}{\sigma + \epsilon}
\end{equation}
where $\mu$ and $\sigma$ denote the mean and standard deviation of the group rewards $\{R(\tau_i)\}_{i=1}^G$, and $\epsilon$ is a small constant ensuring numerical stability. Excluding entropy and KL regularization terms for simplicity, GRPO optimizes the expected trajectory advantage as follows:
\begin{equation}
    \label{eq:grpo_objective}
    \small
    \mathcal{J}_{\mathrm{GRPO}}(\theta) = \mathbb{E}_{\,\mathbf{x},\;\{\tau_i\}_{i=1}^{G}\sim\pi_{\theta_{\mathrm{old}}}} \left[ \frac{1}{G}\sum_{i=1}^{G}\frac{1}{|\tau_i|}\sum_{t=1}^{|\tau_i|} \min\!\left( \rho_{i,t}(\theta) \textcolor{red}{{A}_i}, \operatorname{clip}\big(\rho_{i,t}(\theta), 1-\varepsilon, 1+\varepsilon\big)\,\textcolor{red}{{A}_i} \right) \right]
\end{equation}
where $|\tau_i|$ is the trajectory length (in tokens), and $y_{i,t}$ denotes the $t$-th generated token within $\tau_i$. The importance ratio $\rho_{i,t}(\theta) = \frac{\pi_{\theta}(y_{i,t} \mid \mathbf{x}, y_{i,<t})}{\pi_{\theta_{\mathrm{old}}}(y_{i,t} \mid \mathbf{x}, y_{i,<t})}$ is used in a clipped surrogate with bounds $[1-\varepsilon, 1+\varepsilon]$ to limit improvements from favorable ratio changes. Differentiating this surrogate gives (see Appx.~\ref{appx:grpo_policy_gradient}):
\begin{equation}
    \label{eq:grpo_policy_gradient}
    \small
    \nabla_\theta \mathcal{J}_{\mathrm{GRPO}}(\theta) = \mathbb{E}_{\mathbf{x},\{\tau_i\}_{i=1}^{G}\sim\pi_{\theta_{\mathrm{old}}}} \left[ \frac{1}{G}\sum_{i=1}^{G}\frac{1}{|\tau_i|}\sum_{t=1}^{|\tau_i|} \underbrace{\textbf{1}^{\mathrm{clip}}_{i,t} \cdot \textcolor{red}{{A}_i}  \cdot  \rho_{i,t}(\theta) \cdot \nabla_\theta \log \pi_\theta(y_{i,t} \mid \mathbf{x}, y_{i,<t})}_{\text{gradient at step } (i,t)}  \right]
\end{equation}
where $\textbf{1}^{\mathrm{clip}}_{i,t}$ indicates whether the surrogate's gradient is active: it is zero when $A_i>0$ and $\rho_{i,t}>1+\varepsilon$, or when $A_i<0$ and $\rho_{i,t}<1-\varepsilon$, and one otherwise, with the boundary convention specified in Appx.~\ref{appx:grpo_policy_gradient}.

\noindent \textbf{Credit Assignment of GRPO.} As shown in Eq.~\ref{eq:grpo_policy_gradient}, the per-step gradient is directly scaled by the credit multiplier $\textbf{1}^{\mathrm{clip}}_{i,t} \cdot \textcolor{red}{{A}_i} \cdot \rho_{i,t}(\theta)$. Under the on-policy setting (where $\rho_{i,t}(\theta) \equiv 1$ and thus $\textbf{1}^{\mathrm{clip}}_{i,t} \equiv 1$), this multiplier strictly reduces to the trajectory-level advantage $\textcolor{red}{{A}_i}$. Consequently, all steps within the same trajectory are indiscriminately updated by the identical scalar, regardless of their individual difference. We term this systematic limitation the \textit{\textbf{dilemma of uniform credit assignment}}.

\section{Methodology}
Since RL fundamentally optimizes policies by incentivizing or suppressing action likelihoods, tailored credit assignment requires understanding both the intrinsic properties of each step (e.g., reasonableness and progress advancement) and how modulating them dictates the final outcome. To quantify these step-level properties, \S~\ref{sec:arg_construction} constructs an Agentic Reference Graph (ARG). Guided by an intervention study probing these modulation effects (\S~\ref{sec:action_intervention_analysis}), \S~\ref{sec:proas_framework} introduces Progress-Aware Advantage Shaping (ProAS), a lightweight framework designed to resolve the uniform credit dilemma via asymmetric advantage shaping.
\begin{figure}[t]
    \centering
    \includegraphics[width=0.85\linewidth]{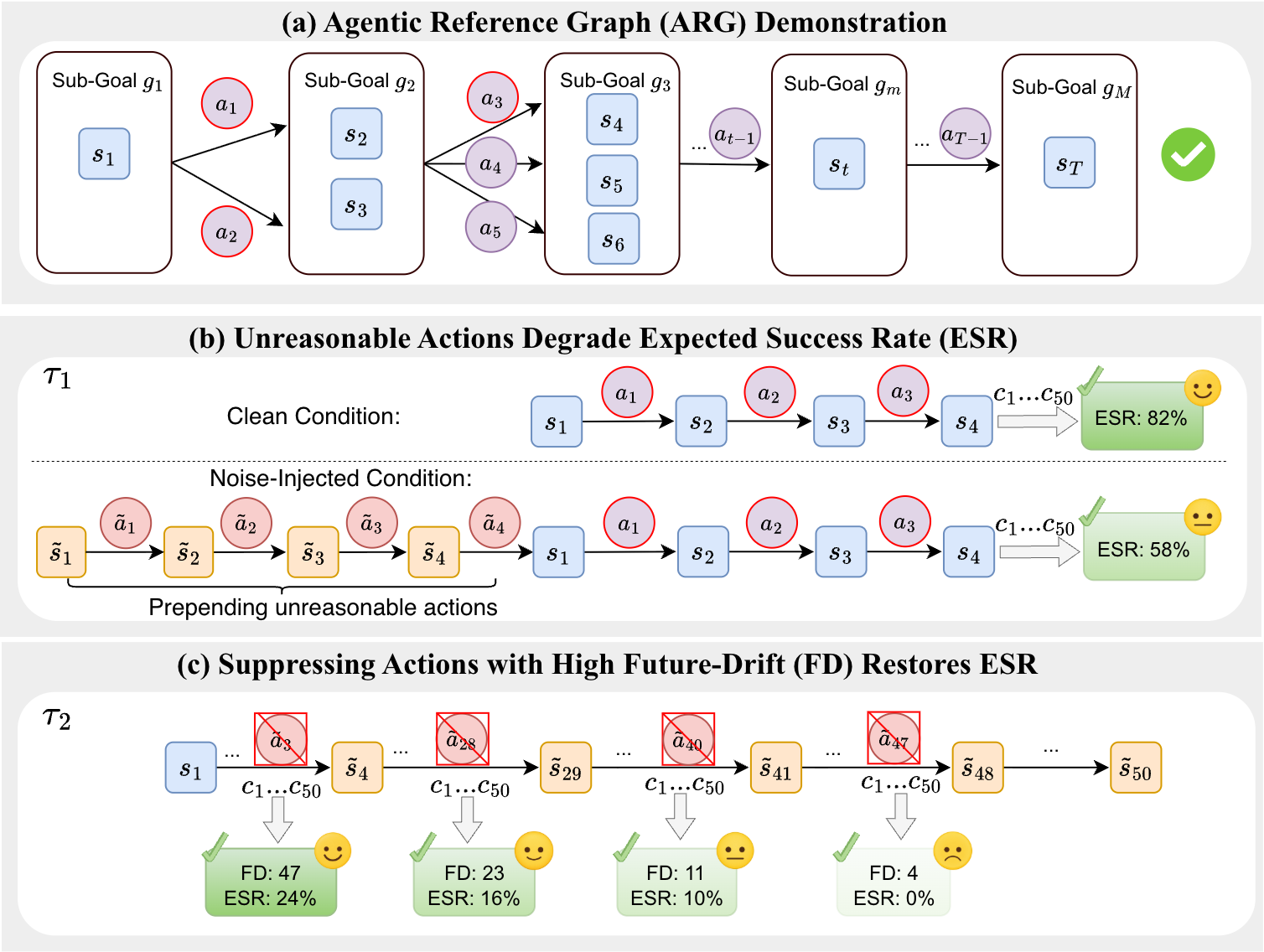}
    \caption{\small
    \textbf{(a) The demonstration of ARG.} Diverse successful trajectories are unified into a DAG that serves as a progress reference.
\textbf{(b) Probing unreasonable actions.} From an identical state, the counterpart injected with unreasonable actions yields a lower Expected Success Rate (ESR) on 50 sampled continuations $\{c_i\}_{i=1}^{50}$.
\textbf{(c) Suppressing by Future-Drift (FD).} Forcibly suppressing the original action at different trajectory points and resampling 50 continuations shows recovered ESR rises with the suppressed action's FD.
    }
    \label{fig:intro_teaser}
\end{figure}


\subsection{Agentic Reference Graph}~\label{sec:arg_construction}
As illustrated in Fig.~\ref{fig:intro_teaser}(a), ARG abstracts the task-solving progression into $M$ ordered sub-goals using pre-collected successful trajectories. Below, we formalize its construction and online evolution, followed by a two-tier annotation scheme to categorize step-level actions.

\noindent \textbf{Construction and Formalization.}
To ensure reliable action assessment, we enforce strict graph purity by pruning invalid transitions and self-loops (example in Appx.~\ref{appx:alfworld_example}). Furthermore, while sub-goals follow a strict sequence, their internal atomic requirements are often permutation-invariant (example in Appx.~\ref{appx:webshop_example}). By modeling each sub-goal's completion as a state equivalence class, we seamlessly unify these diverse execution paths. Formally, the ARG is defined as a Directed Acyclic Graph (DAG) $\mathcal{G} = (\mathcal{V}, \mathcal{E})$:
\begin{definition}[Sub-goals as Vertices]
    \label{def:sub_goals_as_vertices}
    The vertex set $\mathcal{V} = \bigcup_{m=1}^{M} \textbf{g}_{m}$ comprises $M$ sub-goals in sequential order. Each sub-goal $\textbf{g}_m = \{s_{m,1}, s_{m,2}, \dots\}$ is an unordered set of permutation-invariant, valid states, strictly excluding error-induced states and redundant self-loops.
    \end{definition}
    
    \begin{definition}[Actions as Edges]
        \label{def:actions_as_edges}
        The edge set $\mathcal{E} \subseteq \bigcup_{m=1}^{M-1} (\textbf{g}_m \times \mathcal{A}_m \times \textbf{g}_{m+1})$ captures deterministic transitions $(s, a, s')$ between consecutive sub-goals under admissible actions $a \in \mathcal{A}_m$.
        \end{definition}

        \noindent \textbf{Dynamic Evolution.} 
        To capture the diverse spectrum of valid solutions, the ARG evolves dynamically. Upon verifying a novel successful trajectory at the rollout stage of training iteration $k$, we extract its new states ($\Delta\textbf{g}_m$) and actions ($\Delta\mathcal{E}$) to update the current graph $\mathcal{G}^{(k)}$ via strict union:
        \begin{equation}
            \label{eq:arg_evolution}
            \textbf{g}_m^{(k+1)} = \textbf{g}_m^{(k)} \cup \Delta\textbf{g}_m \quad (1 \le m \le M), \quad \text{and} \quad \mathcal{E}^{(k+1)} = \mathcal{E}^{(k)} \cup \Delta\mathcal{E}
        \end{equation}
        These discrete updates strictly adhere to the purity principles, providing an increasingly comprehensive reference for action attribution over time (a raw-to-merged example is provided in Appx.~\ref{appx:alfworld_example}).

\noindent \textbf{Action Attribution.}
We evaluate any rollout trajectory $\tau_{i} = \{(s_t, a_t, s_{t+1})\}_{t=1}^{|\tau_i|}$ generated at iteration $k$ against $\mathcal{G}^{(k)}$ to formally annotate each step $t$ using two binary criteria:

\noindent\textbf{(1) Reasonableness ($\text{Rea}_t$):} Evaluates if an action forms a valid ARG transition: $\text{Rea}_t = \mathbb{I}[ (s_t, a_t, s_{t+1}) \in \mathcal{E}^{(k)} ]$. While it filters explicit errors, it ignores valid but redundant loops.

\noindent\textbf{(2) Progress-Advancement ($\text{PA}_t$):} Strictly quantifies task progression. An action is progress-advancing ($\text{PA}_t = 1$) if it induces a valid ARG state $s_{t+1} \in \mathcal{V}^{(k)}$ while avoiding the historical trajectory $\mathcal{S}_{\le t} = \{s_1, \dots, s_t\}$: $\text{PA}_t = \text{Rea}_t \cdot \mathbb{I}\big[ s_{t+1} \in \mathcal{V}^{(k)} \setminus \mathcal{S}_{\le t} \big]$

Thus, progress-advancing actions are a strict subset of reasonable ones ($\text{PA}_t$ = 1 $\implies$ $\text{Rea}_t$ = 1). Conversely, non-advancing actions ($\text{PA}_t = 0$) encapsulate both explicit errors and structurally valid but stagnant steps.

\subsection{Action-Intervention Analysis}~\label{sec:action_intervention_analysis}
By leveraging the \emph{reasonableness} and \emph{progress-advancement} annotations to assess fine-grained action quality, our pilot study reveals how individual steps dictate outcomes, motivating our solution to the uniform credit assignment dilemma.

\noindent \textbf{Probing Unreasonable Actions.}
Unreasonable actions hinder task progress by polluting the context history and disrupting subsequent decisions~\citep{llm-get-lost-multi-turn}. Notably, they constitute a substantial proportion of successful GRPO trajectories (41.6\% in ALFWorld, 39.6\% in WebShop) of Qwen2.5-1.5B-Instruct, where they are inadvertently reinforced by the uniform credit applied to preferred rollouts. To probe their impact, we select a clean ALFWorld trajectory (details in Appx.~\ref{appx:unreasonable_action_probing}) and sample 50 continuations, $\{c_{i}\}_{i=1}^{50}$, from a fixed state under two context conditions. As illustrated in Fig.~\ref{fig:intro_teaser}(b), the \emph{clean control} retains the original history, whereas the \emph{noise-injected} condition prepends a four-step unreasonable segment. Despite sharing the identical underlying environment state, this visible context noise degrades the Expected Success Rate (ESR) from 82\% to 58\%.

\begin{tcolorbox}[
    colback=blue!3!white,
    colframe=blue!50!black,
    boxrule=0.6pt,
    arc=2mm,
    left=2mm,
    right=2mm,
    top=1.5mm,
    bottom=1.5mm,
    before skip=8pt,
    after skip=8pt
]
    \textbf{Insight 1: } Unreasonable actions in the context actively impede the agent from reaching  success.
\end{tcolorbox}

\noindent \textbf{Suppressing Actions with Varying Future-Drift.}
Conversely, non-advancing (NA) actions are heavily concentrated in failed GRPO trajectories (93.1\% in ALFWorld, 76.1\% in WebShop), where they are suppressed uniformly. Given the autoregressive nature of LLMs, late-stage NA actions are often inevitable consequences of an already polluted context rather than making the mistake independently. Therefore, uniformly suppressing all actions within a failed trajectory is inherently unfair. To quantify the downstream deviation, we define \textbf{Future-Drift} ($\mathcal{D}_{i,t}$, short as FD) as the inclusive count of downstream NA actions from step $t$ in the trajectory $\tau_i$, with $|\tau_i|$  steps:
\begin{equation}
    \label{eq:future_drift}
    \mathcal{D}_{i,t} = \sum_{t'=t}^{|\tau_i|} \big( 1 - \text{PA}_{i,t'} \big)
\end{equation}
To assess whether actions with higher Future-Drift warrant heavier penalization, we perform a targeted suppression intervention on a failed ALFWorld trajectory generated by Qwen2.5-1.5B-Instruct as shown in Fig.~\ref{fig:intro_teaser}(c). By preserving the trajectory prefix up to specific steps, forbidding the original action, and resampling 50 continuations, we measure the recovered ESR. Suppressing the actions with FD values of 47, 23, 11, and 4 yields ESRs of 24\%, 16\%, 10\%, and 0\%, respectively (case details in Appx.~\ref{appx:varying_fd_suppression}). In this case, the higher the suppressed action's Future-Drift, the more likely the agent is to recover and complete the task.
\begin{tcolorbox}[
    colback=blue!3!white,
    colframe=blue!50!black,
    boxrule=0.6pt,
    arc=2mm,
    left=2mm,
    right=2mm,
    top=1.5mm,
    bottom=1.5mm,
    before skip=8pt,
    after skip=8pt
]
    \textbf{Insight 2: } Actions with higher FD should be prioritized for suppression.
\end{tcolorbox}

\subsection{Progress-Aware Advantage Shaping}~\label{sec:proas_framework}
Building upon the empirical insights from \S~\ref{sec:action_intervention_analysis}, we propose the Progress-Aware Advantage Shaping (ProAS) framework, aiming to resolve the uniform credit dilemma by asymmetrically reshaping the trajectory-level advantage $A_i$ into a step-wise, progress-aware signal $\tilde{A}_{i,t}$:
\begin{equation}
    \label{eq:proas_advantage_shaping}
    \tilde{A}_{i,t} = 
    \begin{cases} 
        \text{Rea}_{t} \cdot A_i, & \text{if } A_i > 0, \\ 
        \frac{\mathcal{D}_{i,t}}{\bar{\mathcal{D}}_i + \epsilon} \cdot A_i, & \text{if } A_i < 0 \text{ and } t \in \mathcal{T}, \\ 
        A_i, & \text{otherwise,}
    \end{cases}
\end{equation}
where $\mathcal{T}$ denotes the steps sharing the uniform advantage, $\bar{\mathcal{D}}_i = \frac{1}{\vert\mathcal{T}\vert} \sum_{t' \in \mathcal{T}} \mathcal{D}_{i,t'}$ is the average Future-Drift, and $\epsilon$ ensures numerical stability. This formulation branches into two mechanisms: \textbf{(1) Masking Unreasonable Incentives (MUI):} Aligning with \textbf{Insight 1}, it zeros out the local advantages of invalid actions ($\text{Rea}_t = 0$) in preferred trajectories ($A_i > 0$, usually successful) to prevent inadvertent reinforcement. \textbf{(2) Asymmetric Penalty Redistribution (APR):} Reflecting \textbf{Insight 2}, it leverages normalized Future-Drift to assign proportionally heavier penalties to early pivotal actions in  unpreferred trajectories ($A_i < 0$, usually failed), rectifying the unfairness of uniform suppression.

\section{Experimental Setup} \label{sec:experimental_setup}
To empirically validate the proposed ProAS framework, we conduct extensive experiments across complex, long-horizon agentic environments. Our experiments are structured to answer the following core research questions: 

\noindent \textbf{RQ1: Performance \& Efficiency (\S~\ref{sec:experimental_results}).} Can ProAS consistently elevate task success rates and accelerate training convergence across different model scales under different environments? 

\noindent \textbf{RQ2: Algorithm Compatibility (\S~\ref{sec:experimental_results}).} Does ProAS seamlessly integrate with mainstream group-based RL baselines (e.g., GRPO, DAPO, GiGPO) as a robust plug-and-play module?

\noindent \textbf{RQ3: Advantage Shaping Design (\S~\ref{sec:adv_shap_ablation}).} Are MUI and APR each individually necessary and mutually synergistic, and does ProAS's Future-Drift-based redistribution outperform a simpler masking heuristic?

\noindent \textbf{RQ4: Robustness to ARG (\S~\ref{sec:arg_study}).} Does ProAS remain effective when the reference ARG is randomized, incomplete, static (non-evolving), or distilled from a teacher model?

\subsection{Models and Data}
\noindent \textbf{Models.}
Our experiments are conducted across two different model scales: Qwen2.5-1.5B-Instruct and Qwen2.5-7B-Instruct~\citep{qwen25technicalreport}, which are denoted as 1.5B and 7B for brevity.

\noindent \textbf{Data.}
We train and validate on ALFWorld~\citep{shridhar2021alfworld} and WebShop~\citep{yao2023webshopscalablerealworldweb}, two classical long-horizon multi-turn agentic environments (details in Appx.~\ref{appx:environment_details}). ALFWorld is a text-based embodied benchmark over six household task categories (\texttt{Pick}, \texttt{Look}, \texttt{Clean}, \texttt{Heat}, \texttt{Cool}, \texttt{Pick2}, details in Table.~\ref{tab:alfworld-task-family-distribution}). Our 150-step training comprises 1777/1767/1784 unique executable tasks of seed 0/1/2 for training and 128 for validation. WebShop (Table~\ref{tab:webshop-task-distribution}) is an HTML-based online shopping environment, with 2015/2011/2036 tasks of seed 0/1/2 used for training and 128 for validation.

\subsubsection{Evaluation Metrics}~\label{sec:metric}
\noindent \textbf{Performance Metrics.}
For ALFWorld, we report the average Success Rate (SR) across the entire validation set, alongside the per-category average SR. For WebShop, we report both the overall average SR and the Task Score (TS). TS is a checklist metric on a scale of 0 to 100: it averages the satisfaction of the requested attributes, options, and price constraint, and then scales this score by how well the purchased product matches the requested product type.

\noindent \textbf{Efficiency Metrics.}
We quantify training efficiency via speedup $\text{Spd@}K = N_{\text{base} \to K} / N_{\text{ProAS} \to K}$, which is defined as the ratio of RL training steps required by the baseline versus ProAS to first reach a target $K$ of SR. Reflecting task difficulties, we set $K=70\%$ for ALFWorld and $K=50\%$ for WebShop.


\subsubsection{Implementation Details}
We conduct experiments upon Verl-Agent~\citep{feng2025group}, training all models for 150 RL steps. We validate the policy every 10 RL steps and report metrics from the checkpoint achieving the highest SR. Unless otherwise specified, all experiments are conducted across three random seeds (0, 1, and 2), reporting the mean and sample standard deviation (std). See details in Appx.~\ref{appx:shared_grpo_hyperparameters}.

\noindent \textbf{ARG Initialization.}
Unless otherwise specified, we initialize the ARG in ALFWorld using its official crowdsourced annotations. Given the strictly sequential nature of tasks in ALFWorld, we formulate each individual step as a distinct sub-goal. For WebShop, the ARG is initialized using its officially annotated sub-goals alongside their corresponding states and actions.

\subsubsection{Baselines}
We consider baselines of different paradigms: (1) Proprietary models: GPT-5.6-Sol~\citep{openai2026gpt56}, the state-of-the-art model from OpenAI. We apply the reasoning effort as high, and evaluate it with ReAct prompting~\citep{yao2022react}. (2) Prompting: Qwen2.5-1.5B-Instruct and Qwen2.5-7B-Instruct, the starting point to be trained. (3) Supervised Fine-Tuning (SFT): Tuning the agent by imitating curated turn-level oracle trajectories from the same training tasks as RL baselines (Details in Appx.~\ref{appx:sft_details}). (4) RL: the classical actor-crtic algorithm PPO~\citep{ppo}; mainstream group-based RL algorithms GRPO~\citep{shao2024deepseekmathpushinglimitsmathematical}, 
DAPO~\citep{yu2025dapoopensourcellmreinforcement}, GiGPO~\citep{feng2025group}; the state-of-the-art (SOTA) agentic RL method MiGPO~\citep{MiGPO} with segment-level credit assignment.

\section{Results and Analysis}
\label{sec:experimental_results}

\subsection{Main Results}
Table~\ref{tab:alfworld_webshop} summarizes the main evaluation results on ALFWorld and WebShop. We structure our analysis to directly address our research questions regarding performance, efficiency (RQ1), and algorithm compatibility (RQ2).

\noindent \textbf{Baseline Comparisons.}
We first contrast ProAS against established learning paradigms. While both SFT and ProAS leverage expert annotations as supervisory signals, ProAS abstracts them into ARGs to structurally guide RL, significantly outperforming SFT's direct behavior imitation. Specifically, GiGPO+ProAS achieves 91.9\% and 94.3\% SR on the 1.5B and 7B ALFWorld benchmarks, respectively, surpassing SFT by 12.2 and 14.4 points. Furthermore, ProAS consistently eclipses PPO, a conventional critic-based RL algorithm, across all settings. ProAS variants outpace PPO by 5.5 to 18.7 points in ALFWorld SR, alongside steady improvements of 4.1 to 12.2 points in WebShop TS and 3.4 to 11.5 points in WebShop SR. Ultimately, these results confirm that explicitly integrating ARG-derived progress signals into RL is vastly superior to both passive trajectory imitation and conventional critic-based credit assignment.

\begingroup
\newcommand{\avgstd}[2]{#1\textsubscript{\tiny$\pm$#2}}
\begin{table*}[!t]
    \centering
    \caption{Main results on ALFWorld and WebShop. ALFWorld reports success
    rates (\%) for six task types and all tasks; WebShop reports task score and
    success rate (\%). Boldface indicates that the ProAS framework improves over
    its corresponding base RL algorithm, while a superscript $*$ marks the
    best value within the corresponding model-size block. Results are the mean
    and sample standard deviation over three seeds.}
    \label{tab:alfworld_webshop}
    \scriptsize
    \renewcommand{\arraystretch}{1.04}
    \setlength{\tabcolsep}{2.7pt}
    \resizebox{\textwidth}{!}{%
    \begin{tabular}{lcccccccc|ccc}
    \toprule
    \multirow{2}{*}{Method}
    & \multicolumn{8}{c|}{\textbf{ALFWorld}}
    & \multicolumn{3}{c}{\textbf{WebShop}} \\
    \cmidrule(lr){2-9} \cmidrule(lr){10-12}
    & Pick & Look & Clean & Heat & Cool & Pick2 & ALL
    & Spd.@70 & TS & SR & Spd.@50 \\
    \midrule
    \multicolumn{12}{l}{\textit{Proprietary Model}} \\
    GPT-5.6-Sol (ReAct)
    & 90.9 & 83.3 & 68.0 & 83.3 & 79.2 & 68.2 & 78.9
    & -- & 27.6 & 21.9 & -- \\
    \midrule
    \multicolumn{12}{l}{\textit{Qwen2.5-1.5B-Instruct}} \\
    Direct Prompt
    & \avgstd{24.2}{0.0} & \avgstd{25.0}{0.0} & \avgstd{4.0}{0.0}
    & \avgstd{8.3}{0.0} & \avgstd{0.0}{0.0} & \avgstd{4.5}{0.0}
    & \avgstd{10.9}{0.0}
    & -- & \avgstd{25.3}{0.0} & \avgstd{4.7}{0.0} & -- \\
    \addlinespace[1pt]
    SFT
    & \avgstd{99.0}{1.7}\textsuperscript{*} & \avgstd{91.7}{8.3} & \avgstd{94.7}{2.3}
    & \avgstd{100.0}{0.0}\textsuperscript{*} & \avgstd{93.1}{2.4}\textsuperscript{*} & \avgstd{1.5}{2.6}
    & \avgstd{79.7}{0.8}
    & -- & \avgstd{78.2}{1.7} & \avgstd{64.3}{3.0} & -- \\
    \addlinespace[1pt]
    PPO
    & \avgstd{82.8}{7.6} & \avgstd{58.3}{8.3} & \avgstd{80.0}{4.0}
    & \avgstd{75.0}{8.3} & \avgstd{76.4}{6.4} & \avgstd{54.5}{9.1}
    & \avgstd{73.2}{4.3}
    & -- & \avgstd{75.3}{11.1} & \avgstd{55.2}{6.6} & -- \\
    \addlinespace[1pt]
    MiGPO
    & \avgstd{97.0}{3.0} & \avgstd{91.7}{0.0} & \avgstd{98.7}{2.3}\textsuperscript{*}
    & \avgstd{100.0}{0.0}\textsuperscript{*} & \avgstd{84.7}{6.4} & \avgstd{75.8}{2.6}
    & \avgstd{91.1}{1.6}
    & -- & \avgstd{79.5}{1.2} & \avgstd{53.6}{1.8} & -- \\
    \addlinespace[1pt]
    GRPO
    & \avgstd{80.8}{4.6} & \avgstd{69.4}{4.8} & \avgstd{70.7}{6.1}
    & \avgstd{75.0}{8.3} & \avgstd{63.9}{9.6} & \avgstd{59.1}{9.1}
    & \avgstd{70.3}{2.8}
    & -- & \avgstd{82.4}{1.3} & \avgstd{57.0}{3.1} & -- \\
    \rowcolor{blue!8}GRPO+ProAS
    & \textbf{\avgstd{90.9}{5.2}} & \textbf{\avgstd{86.1}{4.8}} & \textbf{\avgstd{90.7}{2.3}}
    & \textbf{\avgstd{91.7}{8.3}} & \textbf{\avgstd{83.3}{4.2}} & \textbf{\avgstd{78.8}{5.2}}
    & \textbf{\avgstd{87.0}{2.3}}
    & $1.59\times$ & \textbf{\avgstd{85.3}{2.4}} & \textbf{\avgstd{64.6}{1.2}} & $1.79\times$\textsuperscript{*} \\
    \addlinespace[1pt]
    DAPO
    & \avgstd{89.9}{1.7} & \avgstd{83.3}{0.0} & \avgstd{89.3}{12.2}
    & \avgstd{88.9}{4.8} & \avgstd{75.0}{8.3} & \avgstd{66.7}{2.6}
    & \avgstd{82.3}{5.0}
    & -- & \avgstd{79.2}{2.5} & \avgstd{57.3}{4.6} & -- \\
    \rowcolor{blue!8}DAPO+ProAS
    & \textbf{\avgstd{93.9}{3.0}} & \textbf{\avgstd{86.1}{9.6}} & \avgstd{84.0}{6.9}
    & \textbf{\avgstd{94.4}{4.8}} & \textbf{\avgstd{83.3}{4.2}} & \textbf{\avgstd{77.3}{4.5}}
    & \textbf{\avgstd{86.5}{1.2}}
    & $1.27\times$ & \textbf{\avgstd{87.2}{0.7}} & \textbf{\avgstd{66.7}{1.2}}\textsuperscript{*} & $1.61\times$ \\
    \addlinespace[1pt]
    GiGPO
    & \avgstd{91.9}{1.7} & \avgstd{83.3}{0.0} & \avgstd{94.7}{4.6}
    & \avgstd{94.4}{4.8} & \avgstd{84.7}{2.4} & \avgstd{89.4}{2.6}\textsuperscript{*}
    & \avgstd{90.1}{0.5}
    & -- & \avgstd{83.9}{0.4} & \avgstd{64.3}{3.9} & -- \\
    \rowcolor{blue!8}GiGPO+ProAS
    & \textbf{\avgstd{94.9}{1.7}} & \textbf{\avgstd{94.4}{4.8}}\textsuperscript{*} & \avgstd{93.3}{2.3}
    & \textbf{\avgstd{97.2}{4.8}} & \textbf{\avgstd{88.9}{2.4}} & \avgstd{84.8}{2.6}
    & \textbf{\avgstd{91.9}{0.5}}\textsuperscript{*}
    & $1.61\times$\textsuperscript{*} & \textbf{\avgstd{87.5}{1.7}}\textsuperscript{*} & \textbf{\avgstd{66.7}{3.2}}\textsuperscript{*} & $1.12\times$ \\
    \addlinespace[1pt]
    \midrule
    \multicolumn{12}{l}{\textit{Qwen2.5-7B-Instruct}} \\
    Direct Prompt
    & \avgstd{35.4}{3.5} & \avgstd{33.3}{8.3} & \avgstd{13.3}{6.1}
    & \avgstd{5.6}{4.8} & \avgstd{9.7}{2.4} & \avgstd{3.0}{5.2}
    & \avgstd{17.7}{0.9}
    & -- & \avgstd{3.4}{1.3} & \avgstd{0.5}{0.5} & -- \\
    \addlinespace[1pt]
    SFT
    & \avgstd{100.0}{0.0}\textsuperscript{*} & \avgstd{88.9}{4.8} & \avgstd{96.0}{4.0}
    & \avgstd{100.0}{0.0}\textsuperscript{*} & \avgstd{93.1}{2.4}\textsuperscript{*} & \avgstd{1.5}{2.6}
    & \avgstd{79.9}{0.5}
    & -- & \avgstd{81.9}{0.7} & \avgstd{70.1}{0.9} & -- \\
    \addlinespace[1pt]
    PPO
    & \avgstd{97.0}{3.0} & \avgstd{72.2}{4.8} & \avgstd{89.3}{8.3}
    & \avgstd{94.4}{4.8} & \avgstd{77.8}{6.4} & \avgstd{63.6}{12.0}
    & \avgstd{83.6}{4.1}
    & -- & \avgstd{84.1}{4.1} & \avgstd{66.7}{4.3} & -- \\
    \addlinespace[1pt]
    MiGPO
    & \avgstd{100.0}{0.0}\textsuperscript{*} & \avgstd{94.4}{4.8} & \avgstd{96.0}{4.0}
    & \avgstd{80.6}{4.8} & \avgstd{84.7}{2.4} & \avgstd{87.9}{2.6}
    & \avgstd{91.9}{1.2}
    & -- & \avgstd{88.0}{2.2} & \avgstd{67.2}{3.1} & -- \\
    \addlinespace[1pt]
    GRPO
    & \avgstd{90.9}{0.0} & \avgstd{72.2}{9.6} & \avgstd{92.0}{6.9}
    & \avgstd{77.8}{17.3} & \avgstd{76.4}{4.8} & \avgstd{63.6}{4.5}
    & \avgstd{80.7}{3.9}
    & -- & \avgstd{85.6}{5.2} & \avgstd{67.7}{5.5} & -- \\
    \rowcolor{blue!8}GRPO+ProAS
    & \textbf{\avgstd{97.0}{3.0}} & \textbf{\avgstd{91.7}{8.3}} & \textbf{\avgstd{94.7}{2.3}}
    & \textbf{\avgstd{86.1}{4.8}} & \textbf{\avgstd{77.8}{2.4}} & \textbf{\avgstd{83.3}{6.9}}
    & \textbf{\avgstd{89.1}{2.3}}
    & $2.20\times$ & \textbf{\avgstd{89.0}{0.6}} & \textbf{\avgstd{70.8}{0.9}}\textsuperscript{*} & $1.91\times$\textsuperscript{*} \\
    \addlinespace[1pt]
    DAPO
    & \avgstd{97.0}{3.0} & \avgstd{83.3}{14.4} & \avgstd{96.0}{4.0}
    & \avgstd{88.9}{9.6} & \avgstd{76.4}{6.4} & \avgstd{84.8}{6.9}
    & \avgstd{88.8}{5.2}
    & -- & \avgstd{87.1}{3.6} & \avgstd{68.8}{4.7} & -- \\
    \rowcolor{blue!8}DAPO+ProAS
    & \avgstd{97.0}{3.0} & \textbf{\avgstd{86.1}{12.7}} & \textbf{\avgstd{97.3}{4.6}}
    & \avgstd{88.9}{9.6} & \textbf{\avgstd{83.3}{8.3}} & \avgstd{78.8}{5.2}
    & \textbf{\avgstd{89.6}{2.0}}
    & $2.05\times$ & \textbf{\avgstd{88.2}{2.8}} & \textbf{\avgstd{70.6}{0.9}} & $1.62\times$ \\
    \addlinespace[1pt]
    GiGPO
    & \avgstd{97.0}{3.0} & \avgstd{100.0}{0.0}\textsuperscript{*} & \avgstd{97.3}{2.3}
    & \avgstd{97.2}{4.8} & \avgstd{87.5}{0.0} & \avgstd{86.4}{9.1}
    & \avgstd{93.8}{2.1}
    & -- & \avgstd{88.3}{1.8} & \avgstd{70.3}{1.6} & -- \\
    \rowcolor{blue!8}GiGPO+ProAS
    & \textbf{\avgstd{99.0}{1.7}} & \avgstd{88.9}{4.8} & \textbf{\avgstd{98.7}{2.3}}\textsuperscript{*}
    & \avgstd{97.2}{4.8} & \avgstd{86.1}{10.5} & \textbf{\avgstd{92.4}{2.6}}\textsuperscript{*}
    & \textbf{\avgstd{94.3}{2.4}}\textsuperscript{*}
    & $2.44\times$\textsuperscript{*} & \textbf{\avgstd{89.1}{2.7}}\textsuperscript{*} & \avgstd{70.1}{2.3} & $1.58\times$ \\
    \bottomrule
    \end{tabular}%
    }
\end{table*}
\endgroup

\noindent \textbf{State-of-the-Art Performance.}
Equipped with ProAS, group-based RL algorithms consistently establish new SOTA records. In ALFWorld, GiGPO+ProAS attains the highest SR of 94.3\% on the 7B model. This surpasses the strong segment-level baseline MiGPO and eclipses the proprietary GPT-5.6-Sol by a massive 15.4 points. WebShop presents a notably steeper challenge, evidenced by GPT-5.6-Sol's severe struggle at 21.9\% SR. Despite this difficulty, ProAS variants dominate the leaderboard. On the 1.5B model, both DAPO and GiGPO integrated with ProAS tie for the best SR at 66.7\%, outperforming MiGPO by a striking 13.1 points. On the 7B scale, GRPO+ProAS peaks at 70.8\%, further extending the lead over MiGPO.

\noindent \textbf{Algorithm Compatibility.}
Crucially, ProAS demonstrates remarkable plug-and-play compatibility, universally uplifting baseline RL algorithms across all tested configurations. For ALFWorld, equipping GRPO with ProAS yields the most dramatic improvements, boosting the SR by 16.7 points on the 1.5B model and 8.4 points on the 7B model. Consistent gains are mirrored in WebShop, where DAPO+ProAS generates a 9.4-point uplift at the 1.5B scale, and GRPO+ProAS provides a 3.1-point boost at 7B. This confirms ProAS as a versatile enhancer that elevates both the performance floor and ceiling of diverse group-based RL paradigms.

\noindent \textbf{Training Efficiency.}
Beyond asymptotic performance, ProAS significantly accelerates the training process. On average, it speeds up convergence to target success rates by approximately $1.5\times$ for the 1.5B model across both environments. Impressively, this efficiency gain scales favorably with model size: on the 7B model, ProAS accelerates learning by an average of $2.2\times$ on ALFWorld and $1.7\times$ on WebShop.

\subsection{Ablation Study}
\label{sec:ablation_studies}
We ablate the core components of ProAS on the 1.5B model using GRPO as the base algorithm. \S~\ref{sec:adv_shap_ablation} ablates our advantage shaping mechanisms, demonstrating that disabling any single technique lowers the overall success rate and disrupts their synergy. Subsequently, \S~\ref{sec:arg_study} relaxes the ARG construction constraints to investigate ProAS's robustness against reference quality and origin, while highlighting the critical necessity of the online dynamic evolution mechanism.

\subsubsection{Advantage Shaping Ablation} 
\label{sec:adv_shap_ablation}
To isolate the contributions of our specific advantage shaping interventions, we evaluate five ProAS variants:  PR (Progress-aware Reward), ProAS$_{-}$, ProAS$_{+}$,  ProAS$_{\text{msk}\pm}$ and ProAS$_{\text{suffix}}$, as shown in Table~\ref{tab:ablation_study}(a).

First, we contrast advantage shaping against reward shaping. {PR} incorporates ARG progress annotations directly as an auxiliary step-level reward term rather than reshaping the advantage, as Eq.~\ref{eq:progress_reward}. PR underperforms ProAS by 14.1 and 3.4 SR points on ALFWorld and WebShop, respectively. \textbf{This confirms that explicitly modulating trajectory-level advantages is a superior mechanism for injecting structural progress signals compared to the denser reward design in our studies.}


Next, we ablate individual core components, MUI and APR. {ProAS$_{-}$} disables MUI on preferred trajectories ($A_i > 0$), incurring Success Rate (SR) drops of 15.4 and 0.8 points on ALFWorld and WebShop, respectively; {ProAS$_{+}$} disables APR on unpreferred ones ($A_i < 0$), causing a 3.7-point SR drop on ALFWorld and a 1.2-point Task Score (TS) drop on WebShop. These results reveal complementary roles for MUI and APR: MUI prevents reinforcement of unreasonable actions in positively advantaged trajectories, whereas APR differentiates penalties for actions with different Future-Drift in negatively advantaged trajectories. Although their individual benefits vary by environment and metric, \textbf{both components contribute to ProAS’s overall effectiveness and are jointly essential to its design}.

Finally, we evaluate two simpler heuristics for negative trajectories ($A_i < 0$). \textbf{ProAS$_{\text{msk}\pm}$} preserves MUI but replaces APR by naively masking reasonable actions ($\text{Rea}_t = 1$) to avoid their suppression, which underperforms ProAS by 14.1 and 0.8 SR points on ALFWorld and WebShop. \textbf{ProAS$_{\text{suffix}}$} replaces the Future-Drift $\mathcal{D}_{i,t}$ in Eq.\ref{eq:proas_advantage_shaping} with a naive suffix length ($\mathcal{D}_{i,t}^{\mathrm{suffix}} = |\tau_{i}| - t + 1$), trailing ProAS by 5.5 and 1.6 SR points. \textbf{Thus, Future-Drift goes beyond the naive masking and simple positional heuristics.} It quantifies each step’s relative responsibility for the penalty, namely its negative credit, and redistributes the penalty accordingly, which drives the superior performance.


Notably, all three variants still outperform native GRPO on the aggregate metrics, highlighting \textbf{the fundamental necessity of leveraging progress signals to resolve the uniform credit assignment dilemma.}

\subsubsection{ARG Ablation} 
\label{sec:arg_study}
\begingroup
\newcommand{\avgstd}[2]{#1\textsubscript{\tiny$\pm$#2}}
\begin{table*}[!t]
    \centering
    \caption{Ablation study with Qwen2.5-1.5B-Instruct. ALFWorld reports
    success rates (SR, \%) for six task types and all tasks; WebShop reports task
    score (TS, \%) and success rate (SR, \%). Training results are the mean and sample
    standard deviation over three seeds. \textbf{Values higher than the GRPO
    baseline are shown in bold; a superscript $*$ marks the best mean in each
    column.}}
    \label{tab:ablation_study}
    \scriptsize
    \renewcommand{\arraystretch}{1.04}
    \setlength{\tabcolsep}{2.7pt}
    \resizebox{\textwidth}{!}{%
    \begin{tabular}{lccccccc|cc}
    \toprule
    \multirow{2}{*}{Method}
    & \multicolumn{7}{c|}{\textbf{ALFWorld}}
    & \multicolumn{2}{c}{\textbf{WebShop}} \\
    \cmidrule(lr){2-8} \cmidrule(lr){9-10}
    & Pick & Look & Clean & Heat & Cool & Pick2 & ALL
    & TS & SR \\
    \midrule
    GRPO
    & \avgstd{80.8}{4.6} & \avgstd{69.4}{4.8} & \avgstd{70.7}{6.1}
    & \avgstd{75.0}{8.3} & \avgstd{63.9}{9.6} & \avgstd{59.1}{9.1}
    & \avgstd{70.3}{2.8}
    & \avgstd{82.4}{1.3} & \avgstd{57.0}{3.1} \\
    \rowcolor{blue!8}GRPO + ProAS
    & \textbf{\avgstd{90.9}{5.2}}\textsuperscript{*} & \textbf{\avgstd{86.1}{4.8}}\textsuperscript{*}
    & \textbf{\avgstd{90.7}{2.3}} & \textbf{\avgstd{91.7}{8.3}}\textsuperscript{*}
    & \textbf{\avgstd{83.3}{4.2}}\textsuperscript{*} & \textbf{\avgstd{78.8}{5.2}}
    & \textbf{\avgstd{87.0}{2.3}}\textsuperscript{*}
    & \textbf{\avgstd{85.3}{2.4}} & \textbf{\avgstd{64.6}{1.2}} \\
    \midrule
    \multicolumn{10}{l}{\textit{(a) Advantage Shaping Ablation}} \\
    GRPO + PR
    & \textbf{\avgstd{85.9}{4.6}} & \avgstd{44.4}{4.8} & \textbf{\avgstd{86.7}{4.6}}
    & \textbf{\avgstd{83.3}{8.3}} & \textbf{\avgstd{77.8}{6.4}} & \avgstd{42.4}{5.2}
    & \textbf{\avgstd{72.9}{2.3}}
    & \textbf{\avgstd{83.8}{2.6}} & \textbf{\avgstd{61.2}{2.5}} \\
    GRPO + ProAS$_{-}$
    & \textbf{\avgstd{83.8}{9.7}} & \avgstd{52.8}{19.2} & \textbf{\avgstd{78.7}{8.3}}
    & \textbf{\avgstd{77.8}{9.6}} & \avgstd{63.9}{19.2} & \textbf{\avgstd{60.6}{5.2}}
    & \textbf{\avgstd{71.6}{10.4}}
    & \textbf{\avgstd{86.0}{2.4}}\textsuperscript{*} & \textbf{\avgstd{63.8}{5.9}} \\
    GRPO + ProAS$_{+}$
    & \textbf{\avgstd{86.9}{1.7}} & \textbf{\avgstd{83.3}{8.3}} & \textbf{\avgstd{85.3}{6.1}}
    & \textbf{\avgstd{88.9}{12.7}} & \textbf{\avgstd{76.4}{2.4}} & \textbf{\avgstd{80.3}{2.6}}\textsuperscript{*}
    & \textbf{\avgstd{83.3}{1.8}}
    & \textbf{\avgstd{84.1}{2.1}} & \textbf{\avgstd{64.6}{1.8}} \\
    GRPO + ProAS$_{\text{msk}\pm}$
    & \avgstd{80.8}{4.6} & \textbf{\avgstd{72.2}{4.8}} & \textbf{\avgstd{74.7}{12.2}}
    & \textbf{\avgstd{83.3}{8.3}} & \textbf{\avgstd{70.8}{4.2}} & \avgstd{56.1}{2.6}
    & \textbf{\avgstd{72.9}{2.4}}
    & \textbf{\avgstd{83.5}{2.0}} & \textbf{\avgstd{63.8}{2.5}} \\
    GRPO + ProAS$_{\text{suffix}}$
    & \textbf{\avgstd{87.9}{3.0}} & \textbf{\avgstd{83.3}{8.3}} & \textbf{\avgstd{85.3}{2.3}}
    & \textbf{\avgstd{88.9}{4.8}} & \textbf{\avgstd{76.4}{6.4}} & \textbf{\avgstd{68.2}{4.5}}
    & \textbf{\avgstd{81.5}{2.0}}
    & \textbf{\avgstd{85.6}{1.1}} & \textbf{\avgstd{63.0}{5.5}} \\
    \midrule
    \multicolumn{10}{l}{\textit{(b) ARG Ablation}} \\
    Random
    & \avgstd{67.7}{7.0} & \avgstd{38.9}{9.6} & \avgstd{56.0}{10.6}
    & \avgstd{44.4}{4.8} & \avgstd{34.7}{19.7} & \avgstd{28.8}{6.9}
    & \avgstd{47.7}{4.8}
    & \avgstd{78.6}{2.1} & \avgstd{49.2}{3.6} \\
    Static
    & \textbf{\avgstd{87.9}{6.1}} & \textbf{\avgstd{83.3}{0.0}} & \textbf{\avgstd{94.7}{2.3}}\textsuperscript{*}
    & \textbf{\avgstd{86.1}{4.8}} & \textbf{\avgstd{70.8}{7.2}} & \textbf{\avgstd{72.7}{4.5}}
    & \textbf{\avgstd{82.8}{2.8}}
    & \textbf{\avgstd{85.2}{1.7}} & \textbf{\avgstd{65.9}{2.7}}\textsuperscript{*} \\
    Incomplete
    & \textbf{\avgstd{85.9}{1.7}} & \textbf{\avgstd{75.0}{8.3}} & \textbf{\avgstd{77.3}{4.6}}
    & \textbf{\avgstd{77.8}{4.8}} & \textbf{\avgstd{76.4}{2.4}} & \textbf{\avgstd{66.7}{10.5}}
    & \textbf{\avgstd{77.3}{2.1}}
    & \textbf{\avgstd{85.0}{2.4}} & \textbf{\avgstd{63.0}{3.2}} \\
    Incomplete \& Static
    & \textbf{\avgstd{82.8}{7.6}} & \textbf{\avgstd{75.0}{8.3}} & \textbf{\avgstd{76.0}{4.0}}
    & \textbf{\avgstd{77.8}{4.8}} & \avgstd{62.5}{12.5} & \avgstd{57.6}{6.9}
    & \textbf{\avgstd{72.1}{5.2}}
    & \avgstd{82.3}{1.4} & \textbf{\avgstd{57.6}{2.4}} \\
    Distill
    & \textbf{\avgstd{85.9}{9.3}} & \textbf{\avgstd{80.5}{4.8}} & \textbf{\avgstd{89.3}{2.3}}
    & \textbf{\avgstd{88.9}{4.8}} & \textbf{\avgstd{80.5}{8.7}} & \textbf{\avgstd{75.7}{2.6}}
    & \textbf{\avgstd{83.6}{0.8}}
    & \textbf{\avgstd{84.8}{1.0}} & \textbf{\avgstd{64.8}{1.4}} \\
    \bottomrule
    \end{tabular}%
    }
\end{table*}
\endgroup

$\texttt{Random}$ replaces ARG-derived annotations with random labels instead of matching against the graph. This causes a catastrophic collapse, dropping SR by 39.3 points on ALFWorld and 15.4 points on WebShop relative to ProAS (Table~\ref{tab:ablation_study}(b)). \textbf{This confirms that clean, ARG-derived annotations are indispensable for accurate credit assignment.}

$\texttt{Incomplete}$ randomly prunes 40\% of the action-state pairs from the clean ARG to simulate severe degradation. Despite this, ProAS still outperforms GRPO by 7.0 points on ALFWorld. This robustness largely comes from ARG evolution (mentioned in Eq.~\ref{eq:arg_evolution}): disabling it (\texttt{Incomplete\&Static}) further drops SR by 5.2 points on ALFWorld and 5.4 points on WebShop relative to \texttt{Incomplete} alone. Evolution matters even with a clean ARG, where disabling it (\texttt{Static}) lowers ALFWorld SR from 87.0\% (GRPO+ProAS) to 82.8\%, though \texttt{Static} still attains the best WebShop SR (65.9\%) among all variants. \textbf{These results confirm the importance of ARG evolution for ProAS, while showing ProAS remains robust to imperfect ARGs.}

$\texttt{Distill}$ tests ProAS with synthetic references, since gold-standard trajectories are often unavailable in practice. We prompt a 7B model to generate successful rollouts across training tasks, building a distilled ARG covering 2,566 tasks. With this synthetic ARG, the 1.5B model reaches 83.6\% SR on ALFWorld and 64.8\% on WebShop, comparable to GRPO+ProAS and 13.3 and 7.8 points above GRPO, respectively. \textbf{This validates ProAS's applicability even on teacher-distilled trajectories.}

\section{Conclusion} \label{sec:conclusion}
\vspace{-0.2cm}
In this work, we expose the \textit{uniform credit assignment dilemma} in critic-free agentic RL, where uniform advantages reinforce contextual noise in preferred trajectories ($A_i > 0$) and ignore the downstream deviation of actions in unpreferred ones ($A_i < 0$). To address this without critic overhead, we proposed Progress-Aware Advantage Shaping (ProAS), which leverages a dynamically evolving Agentic Reference Graph (ARG) to achieve fine-grained credit assignment via Masking Unreasonable Incentives (MUI) and Asymmetric Penalty Redistribution (APR). Extensive evaluations across multiple model scales and environments confirm ProAS as a robust, universal enhancer that accelerates convergence and propels mainstream baselines to new state-of-the-art performance. The ablation study further confirms MUI-APR synergy, necessity of ARG evolution and robustness under incomplete or distilled ARGs. Together, these results establish a highly efficient, scalable pathway for elevating autonomous language agents.

\bibliographystyle{venues/iclr/configuration/iclr_conference}
\bibliography{common/references}

\clearpage
\appendix
\section{Derivation of the GRPO Policy Gradient}
\label{appx:grpo_policy_gradient}

We derive Eq.~\ref{eq:grpo_policy_gradient} from the clipped surrogate in
Eq.~\ref{eq:grpo_objective}, omitting entropy and KL regularization as in the
main text. During each policy update, the sampling policy
$\pi_{\theta_{\mathrm{old}}}$, sampled trajectories, and group-normalized
advantages $A_i$ are fixed; in particular, $\nabla_\theta A_i=0$.
Let $h_{i,t}=(\mathbf{x},y_{i,<t})$ denote the conditioning context and assume
the old policy assigns positive probability to each sampled token.

\paragraph{Clipped token-level surrogate.}
Writing $\rho=\rho_{i,t}(\theta)$, the contribution of token $(i,t)$ is
\begin{equation}
\begin{aligned}
    \ell_{i,t}(\theta)
    &=\min\!\left(\rho A_i,
        \operatorname{clip}(\rho,1-\varepsilon,1+\varepsilon)A_i\right)\\
    &=\begin{cases}
        A_i\min(\rho,1+\varepsilon), & A_i>0,\\
        A_i\max(\rho,1-\varepsilon), & A_i<0,\\
        0, & A_i=0.
    \end{cases}
\end{aligned}
\label{eq:appx_grpo_piecewise}
\end{equation}
Thus, the surrogate is flat only when $A_i>0$ and
$\rho>1+\varepsilon$, or when $A_i<0$ and $\rho<1-\varepsilon$.
A ratio outside the clipping interval does not, by itself, imply zero gradient:
the gradient remains active when the ratio moves in the unfavorable direction.
Away from the clipping boundaries, define
\begin{equation}
    \textbf{1}^{\mathrm{clip}}_{i,t}=
    \begin{cases}
        0, & A_i>0\ \text{and}\ \rho_{i,t}>1+\varepsilon,\\
        0, & A_i<0\ \text{and}\ \rho_{i,t}<1-\varepsilon,\\
        1, & \text{otherwise}.
    \end{cases}
    \label{eq:appx_grpo_clip_indicator}
\end{equation}
At a clipping kink, we adopt the derivative of the active linear branch
(indicator $1$), a valid generalized-gradient convention. For $A_i=0$,
the gradient is zero regardless of the indicator.

\paragraph{Importance-ratio derivative.}
Since the denominator is fixed, the log-derivative identity gives
\begin{equation}
\begin{aligned}
    \nabla_\theta\rho_{i,t}(\theta)
    &=\frac{\nabla_\theta\pi_\theta(y_{i,t}\mid h_{i,t})}
      {\pi_{\theta_{\mathrm{old}}}(y_{i,t}\mid h_{i,t})}\\
    &=\rho_{i,t}(\theta)
      \nabla_\theta\log\pi_\theta(y_{i,t}\mid h_{i,t}).
\end{aligned}
\end{equation}
Applying the chain rule to Eq.~\ref{eq:appx_grpo_piecewise} therefore yields
\begin{equation}
    \nabla_\theta\ell_{i,t}(\theta)
    =\textbf{1}^{\mathrm{clip}}_{i,t}A_i\rho_{i,t}(\theta)
      \nabla_\theta\log\pi_\theta(y_{i,t}\mid h_{i,t}).
\end{equation}

\paragraph{Group-level gradient.}
Under the usual conditions for exchanging differentiation and expectation,
the fixed sampling distribution introduces no additional derivative term.
Averaging the token-level derivatives gives
\begin{equation}
\begin{aligned}
    \nabla_\theta\mathcal{J}_{\mathrm{GRPO}}(\theta)
    &=\mathbb{E}_{\mathbf{x},\{\tau_i\}_{i=1}^{G}\sim\pi_{\theta_{\mathrm{old}}}}
    \Bigg[\frac{1}{G}\sum_{i=1}^{G}\frac{1}{|\tau_i|}
    \sum_{t=1}^{|\tau_i|}
    \textbf{1}^{\mathrm{clip}}_{i,t}A_i\rho_{i,t}(\theta)\\
    &\hspace{35mm}\cdot\nabla_\theta\log\pi_\theta(y_{i,t}\mid h_{i,t})\Bigg],
\end{aligned}
\end{equation}
which recovers Eq.~\ref{eq:grpo_policy_gradient}. At
$\theta=\theta_{\mathrm{old}}$, $\rho_{i,t}=1$ and the clipping indicator is
$1$ for $\varepsilon>0$. Consequently, all tokens in a trajectory share the
same advantage multiplier $A_i$ (and trajectory normalization
$1/|\tau_i|$), although their score-function gradients can differ. This is
the uniform credit-assignment property discussed in Sec.~\ref{sec:grpo}.

\section{Further Implementation Details} \label{appx:shared_grpo_hyperparameters}
\subsection{Training Framework}
For ALFWorld and WebShop, we adopt the Verl-Agent~\citep{feng2025group} framework. To ensure robustness and reproducibility, we sample and freeze three distinct training sets with environment seeds 0, 1, and 2 for independent and repeated training. For fair evaluation, all resulting model checkpoints are assessed on a unified, static validation set of 128 tasks. This validation set is strictly disjoint from the training data, rigorously preventing any potential data leakage.
Validation uses a temperature of $0.4$ in both ALFWorld and WebShop. 

RL and SFT
share the same turn-level input format illustrated in
Tables~\ref{tab:sft_data_alfworld} and~\ref{tab:sft_data_webshop}: each prompt
contains the task instruction, current observation, admissible actions, and the
visible interaction history from the two most recent observation-action turns.
The only difference between SFT and RL data is the assistant response, which is curated offline for SFT
but generated online for RL.

All experiments are conducted on GPUs. For ALFWorld and WebShop, each 1.5B
model run uses 2 GPUs, while each 7B model run uses 4 GPUs. 

\subsection{RL Configuration}
\label{appx:rl_configuration}
\vspace{-0.2cm}
\noindent \textbf{General.}
We follow the standard Verl-Agent~\citep{feng2025group} configuration and use
the same settings across all RL methods unless otherwise specified. Models are
optimized with AdamW using a learning rate of $10^{-6}$, weight decay $0.01$,
and gradient clipping at $1.0$. We use a PPO clipping ratio of $0.2$, an entropy
coefficient of $0.001$, a KL-loss coefficient of $0.01$, and one policy-update
epoch per rollout batch. At each RL iteration, we sample 16 prompts and generate
$G=8$ trajectories per prompt with temperature $1.0$ and top-$p=1.0$.

For ALFWorld, we set the maximum prompt and response lengths to 2,048 and 512
tokens, respectively, cap each episode at 50 steps, and use a PPO mini-batch
size of 256. For WebShop, the corresponding settings are 4,096 prompt tokens,
512 response tokens, 15 steps, and a PPO mini-batch size of 64.

\noindent \textbf{Reward.}
Across all runs, we use the same rule-based sparse outcome reward $R_{\text{o}}$ in both
environments: a successful trajectory receives $10$, whereas an unsuccessful
trajectory receives $0$. The $R_{o}$ is broadcasted into each step in the trajectory.
Each invalid action additionally incurs a penalty of
$-0.1$. Specifically, the response of a step is marked invalid ($\mathbb{I}[\text{invalid}]=1$) if it omits a complete
\thinktag{think}\ldots\thinktag{/think} or
\actiontag{action}\ldots\actiontag{/action} block, or contains Chinese
characters, otherwise, the response is valid ($\mathbb{I}[\text{invalid}]=0$).  Consequently, the reward for step $t^{\prime}$ in trajectory $\tau_i$ is given by:
\begin{equation}
    \label{eq:reward}
    R_{i,t^{\prime}} = R_{\text{o}_{i}} - 0.1 \cdot \mathbb{I}[\text{invalid}_{i,t^{\prime}}]
\end{equation}
For progress-aware reward (PR) mentioned in \S~\ref{sec:adv_shap_ablation}, we add the progress-aware reward term $ R_{\text{p}_{i,t^{\prime}}} = \frac{\sum_{t=1}^{t^{\prime}} \mathrm{PA}_{i,t}}{\sum_{t=1}^{|\tau_i|} \mathrm{PA}_{i,t}}$ for the reward of each step:
\begin{equation}
    \label{eq:progress_reward}
    R_{i,t^{\prime}}
= R_{o_i}
- 0.1 \cdot \mathbb{I}[\mathrm{invalid}_{i,t^{\prime}}]
+ 0.5 \cdot R_{\text{p}_{i,t^{\prime}}}
\end{equation}
This design incorporates the progress-advancing information obtained from ARG into the reward signal.

\noindent \textbf{PPO.}
For PPO~\citep{ppo}, we estimate advantages using generalized
advantage estimation (GAE) with discount factor $\gamma=1.0$
and trace-decay parameter $\lambda=1.0$. We train a separate
critic with AdamW using a learning rate of $10^{-5}$ and
weight decay $0.01$, without a critic-only warmup phase.
Since PPO does not require multiple rollouts per task for
advantage estimation, we sample one rollout per task ($G=1$)
and use a rollout batch of 128 tasks. This matches the
128 sampled trajectories per RL iteration used by the
critic-free methods, which sample 8 rollouts for each of
16 tasks. The policy learning rate, regularization, and
rollout sampling settings follow the shared configuration
above.

\noindent \textbf{DAPO.}
For DAPO~\citep{yu2025dapoopensourcellmreinforcement}, we use asymmetric PPO
clipping with lower and upper ratios of $0.2$ and $0.28$, respectively, and a
dual-clip coefficient of $10.0$. We disable the actor KL loss and entropy
regularization, use token-mean loss aggregation, set the AdamW weight decay to
$0.1$ with 10 learning-rate warmup steps, and disable group filtering.

\noindent \textbf{GiGPO.}
For GiGPO~\citep{feng2025group}, we set the temporal discount factor to
$\gamma=0.95$ and the step-level advantage weight to $1.0$. Step advantages
are normalized by both the group mean and standard deviation in ALFWorld, and
by the group mean only in WebShop.

\noindent \textbf{GiGPO+ProAS.}
Since GiGPO adopts step-level group-normalized advantages, different steps within the same trajectory need not share the same advantage. This, however, does not affect the plug-in integration of ProAS. We apply MUI to actions with excessively positive advantages, and cluster negative-advantage actions within each trajectory according to their individual advantage magnitudes, rescaling them by future-miss normalization to a mean of one per cluster.


\noindent \textbf{MiGPO.}
For MiGPO~\citep{MiGPO}, we use mean-normalized segment-level advantages. In
ALFWorld, we set the milestone-matching threshold to $0.95$, the reverse-decay
factor to $0.9$, and the step-level advantage weight to $10.0$. WebShop uses
environment-provided dynamic milestone signals, with a reverse-decay factor of
$0.95$ and a step-level advantage weight of $1.0$.

\subsection{SFT Details}
\label{appx:sft_details}
\subsubsection{Data Curation.}
To construct a controlled imitation-learning baseline matching the RL methods' task exposure, we curate a supervised dataset for each environment seed (0, 1, 2) from the frozen training tasks:

\noindent \textbf{ALFWorld:} We execute the official annotated expert trajectory to obtain one demonstration per unique training game.

\noindent \textbf{WebShop:} Lacking official experts, we instantiate a deterministic, goal-conditioned oracle from our ARG using task metadata (target product, query, and options). The resulting trajectories strictly follow the ARG's sub-goal progression ($\textbf{g}_1 \rightarrow \dots \rightarrow \textbf{g}_6$, details in Table~\ref{tab:webshop-option-permutation}), explicitly grounding the demonstrations in the same progress structure used by ProAS.

Aligning with the data setting in RL experiments. 
we represent SFT data as turn-level examples rather than complete
trajectories. Each input contains the task instruction, current observation,
admissible actions, and the two most recent interaction turns. Table~\ref{tab:sft_data_alfworld} and Table~\ref{tab:sft_data_webshop} demonstrate the SFT data example of ALFWorld and WebShop.

\addtocounter{table}{2}

\noindent \textbf{Training Configuration.}
We perform full-parameter fine-tuning of Qwen2.5-1.5B-Instruct and
Qwen2.5-7B-Instruct with the SFT trainer based on FSDP in Verl-Agent, minimizing the
token-level negative log-likelihood of the oracle responses. We train for 3
epochs with AdamW ($\beta_1=0.9$, $\beta_2=0.95$), a peak learning rate of
$10^{-5}$, weight decay of $0.01$, and gradient clipping at $1.0$. The learning
rate follows a cosine schedule with a $3\%$ warmup. We use a maximum sequence
length of 4,096 tokens with left truncation and per-GPU micro-batch sizes of 4
and 2 for the 1.5B and 7B models, respectively. We save the checkpoint at the
end of each epoch and evaluate every epoch checkpoint on the same frozen
128-task validation set used for RL. The SFT result is reported from the
checkpoint with the highest validation success rate.

\section{Environment Details} \label{appx:environment_details}
\paragraph{ALFWorld.}
ALFWorld~\citep{shridhar2021alfworld} is a text-based embodied benchmark built
from household tasks in simulated indoor scenes. It covers six task families:
\texttt{single-object manipulation (Pick)}, \texttt{object examination (Look)}, \texttt{cleaning (Clean)}, \texttt{heating (Heat)}, \texttt{cooling (Cool)}, and
\texttt{two-object manipulation (Pick2)}. At each step, the agent observes textual environment
feedback and issues one admissible natural-language action, making ALFWorld a
multi-turn, single-step environment. Table~\ref{tab:alfworld-task-family-distribution}
reports deduplicated task-family counts for the three frozen training sets and the
shared validation set.

\begin{table}[H]
    \centering
    \caption{ALFWorld task-family counts in the frozen training sets and
    validation set. Each training column counts unique game instances within
    one seed. Validation contains 128 distinct tasks shared across seeds.}
    \label{tab:alfworld-task-family-distribution}
    \small
    \begin{tabular*}{\linewidth}{@{\extracolsep{\fill}}ccccc@{}}
        \toprule
        Task family & Training seed 0 & Training seed 1 & Training seed 2 & Validation \\
        \midrule
        Pick         & 391 & 392 & 399 & 33 \\
        Look         & 161 & 162 & 162 & 12 \\
        Clean        & 315 & 311 & 310 & 25 \\
        Heat         & 249 & 241 & 239 & 12 \\
        Cool         & 262 & 268 & 266 & 24 \\
        Pick2        & 399 & 393 & 408 & 22 \\
        \midrule
        Total & 1777 & 1767 & 1784 & 128 \\
        \bottomrule
    \end{tabular*}
\end{table}

\clearpage
\paragraph{WebShop.}
WebShop~\citep{yao2023webshopscalablerealworldweb} is an HTML-based shopping
environment in which the agent must search for a product of \texttt{Fashion}, \texttt{Garden}, \texttt{Electronics}, \texttt{Grocery} and \texttt{Beauty}, then navigate result and
product pages, select required options, and purchase the item that matches the
user instruction. The task naturally involves both ordered milestones, such as
opening the target product before purchasing it, and unordered option choices,
such as selecting size and color in either order. Similar to ALFWorld, WebShop
allows one environment action per interaction turn.
Table~\ref{tab:webshop-task-distribution}
reports the distribution of training sets of seed 0/1/2 and the validation set.

\begin{table}[H]
    \centering
    \caption{WebShop task-category counts in the frozen training sets and
    validation set. Each training column counts unique base tasks within
    one seed. Validation contains 128 distinct tasks shared across seeds.}
    \label{tab:webshop-task-distribution}
    \small
    \begin{tabular*}{\linewidth}{@{\extracolsep{\fill}}ccccc@{}}
        \toprule
        Task category & Training seed 0 & Training seed 1 & Training seed 2 & Validation \\
        \midrule
        Fashion      & 1570 & 1568 & 1577 & 95 \\
        Garden       & 294 & 291 & 283 & 23 \\
        Electronics  & 57 & 53 & 57 & 6 \\
        Grocery      & 37 & 37 & 39 & 2 \\
        Beauty       & 57 & 62 & 80 & 2 \\
        \midrule
        Total & 2015 & 2011 & 2036 & 128 \\
        \bottomrule
    \end{tabular*}
\end{table}

\subsection{Time Analysis}
\begin{figure}[H]
    \centering
    \includegraphics[width=0.85\linewidth]{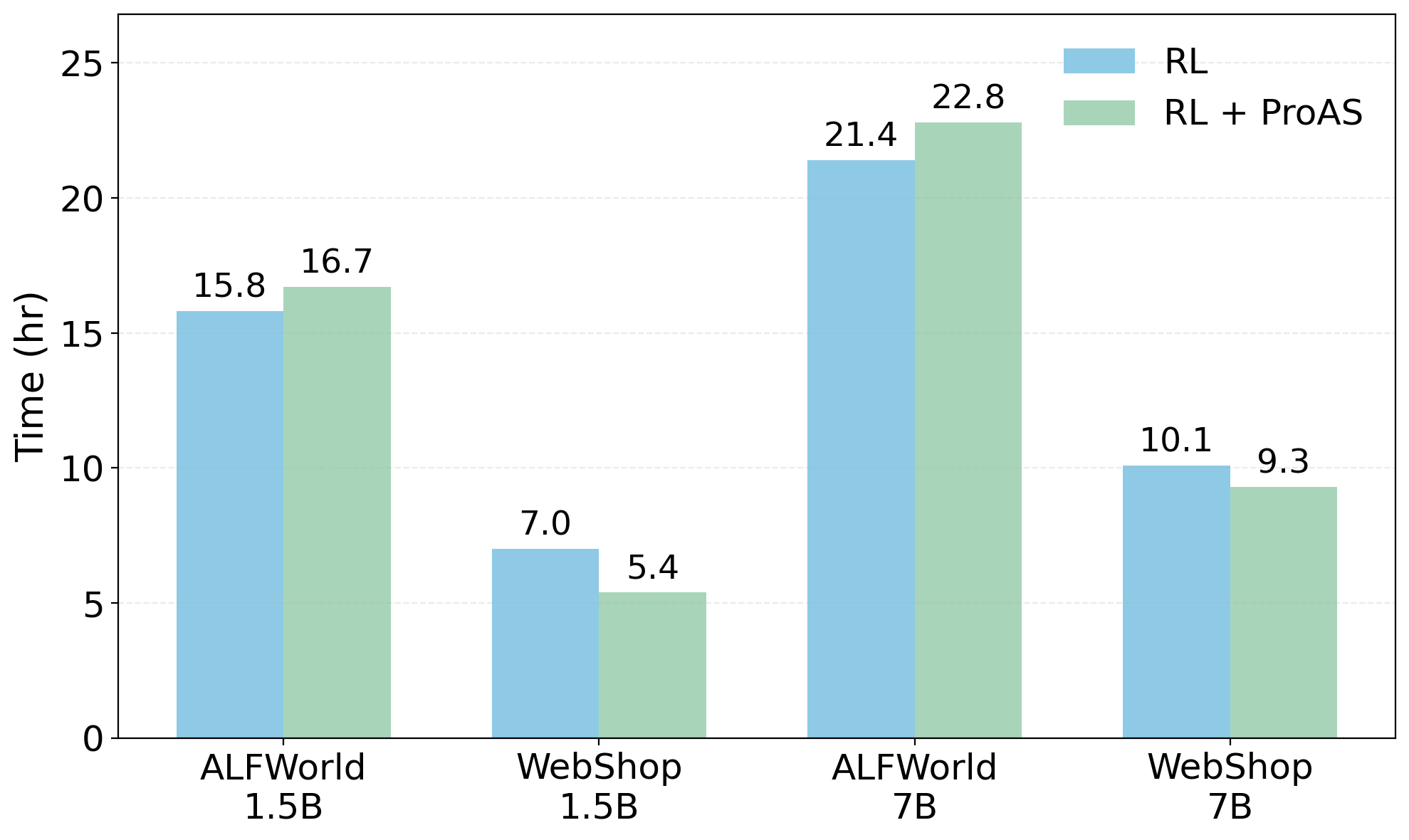}
    \caption{\small
   \textbf{Training time (hr) comparison between vanilla RL and RL + ProAS.} 
   Wall-clock training time across ALFWorld and WebShop with 1.5B and 7B models. 
   Each value is averaged over GRPO, DAPO, and GiGPO runs under seed 1. 
    }
    \label{fig:proas_time_compare}
\end{figure}

We compare the wall-clock training time of vanilla RL and RL + ProAS, measured from TensorBoard logs. As shown in Fig.~\ref{fig:proas_time_compare}, ProAS incurs modest overhead on ALFWorld while achieving lower training time on WebShop. This suggests that ProAS introduces limited computational overhead, as it only requires ARG queries and updates, action-reasonableness checks, and advantage masking and reweighting, without training an additional critic or invoking an extra reward model. The slightly higher training time on ALFWorld may partly reflect its longer interaction horizon, which can accumulate the cost of graph queries, progress assessment, and online graph updates.

\newpage
\section{Case Study}

\subsection{Unreasonable Action Probing Analysis in ALFWorld}
\label{appx:unreasonable_action_probing}

We compare two visible-history conditions while fixing the underlying
environment state and rollout budget. Both conditions replay the same
three-action clean prefix and reach the same continuation point: the agent is at
\texttt{desk 1}, holding \texttt{keychain 2}, with 47 steps remaining. From
this state, we independently sample 50 continuations $\{c_i\}_{i=1}^{50}$ under each condition. In
the noise-injected condition, four unreasonable actions are prepended only to
the policy-visible interaction history; they are not executed in the
continuation environment and therefore do not alter its starting state.

Table~\ref{tab:alfworld-unreasonable-probing} first presents the task ARG and
then traces the clean and noise-injected conditions separately. Only the four
injected unreasonable actions are highlighted in pink. The sampled suffixes are
summarized by a continuation row, followed by the Expected Success Rate (ESR)
for each condition.


The two panels therefore differ only in the pink history segment; their clean
prefix, continuation state, and rollout budget are identical. The observed ESR
decrease from $82\%$ to $58\%$ thus diagnoses the effect of unreasonable content
in the policy-visible history rather than a change in the underlying
environment state.

\subsection{Varying Future-Drift Actions Suppression Analysis in ALFWorld}
\label{appx:varying_fd_suppression}

We examine the suppression of four non-advancing actions from the same 50-step failed ALFWorld
rollout in Table~
\ref{tab:alfworld-varying-fd-suppression} (c), which are actions of steps $3$, $28$, $40$, and
$47$. At each position, we preserve the complete preceding
prefix, reject only an exact match to the original action, and independently
sample 50 continuations $\{c_i\}_{i=1}^{50}$.

The intervention results exhibit a monotonic relation in this case: the ESR is
$24\%$, $16\%$, $10\%$, and $0\%$ when suppressing actions with FD $47$, $23$,
$11$, and $4$, respectively. The larger recovery at higher-FD positions
supports Future-Drift as a diagnostic proxy for an action's downstream failure
burden. Because the four decisions occur at different prefixes and remaining
horizons, this case is intended as direct intervention evidence for the
diagnostic trend rather than a standalone estimate of a universal causal
effect.

\subsection{The ARG Dynamic Evolution in ALFWorld}
\label{appx:alfworld_example}
Table~\ref{tab:alfworld-arg-purity} illustrates the online ARG evolution mechanism using the task \texttt{Look at a pencil under the desk lamp}. Because the environment's success criterion broadly accepts any pencil index (e.g., \texttt{pencil 1} or \texttt{pencil 2}), the task inherently possesses multiple valid solutions that a single static ARG fails to cover. To capture this diversity, any rollout verified as successful during training is first purified by pruning invalid and inadmissible steps. The retained valid path is then seamlessly merged into the current ARG, dynamically expanding its coverage of viable solution trajectories.

\subsection{The Order-Agnostic of Actions in  WebShop}
\label{appx:webshop_example}

Table~\ref{tab:webshop-option-permutation} shows a WebShop case from
Qwen2.5-1.5B-Instruct GRPO rollouts. The task-level ARG is ordered across
sub-goals, but its option-configuration sub-goal $\textbf{g}_3$ contains an
unordered set of required options. A partial $\textbf{g}_3$ state satisfies one
required option, while $\textbf{g}_3$ is completed only after all required
options are selected. The price and attribute constraints, $\textbf{g}_4$ and
$\textbf{g}_5$, are checked at purchase time before the trajectory reaches
$\textbf{g}_6$.

The contrast highlights the intended granularity of action annotations deriving from ARG.
Trajectories A and B differ only within $\textbf{g}_3$: they visit different
intermediate option states but reach the same completion-state equivalence
class, so both option orders are treated as progress-advancing. Trajectory C
shows the complementary cases: a wrong option and a premature purchase are
unsupported by the ARG, whereas a repeated click on an already selected option
can be reasonable but non-advancing. Thus, the ARG preserves ordering
constraints among high-level sub-goals while avoiding spurious penalties for
valid option permutations.

\section{Training Speedup Curves} \label{appx:training_speedup_curves}
\begin{figure*}[!htbp] 
    \centering
    \includegraphics[width=\linewidth]{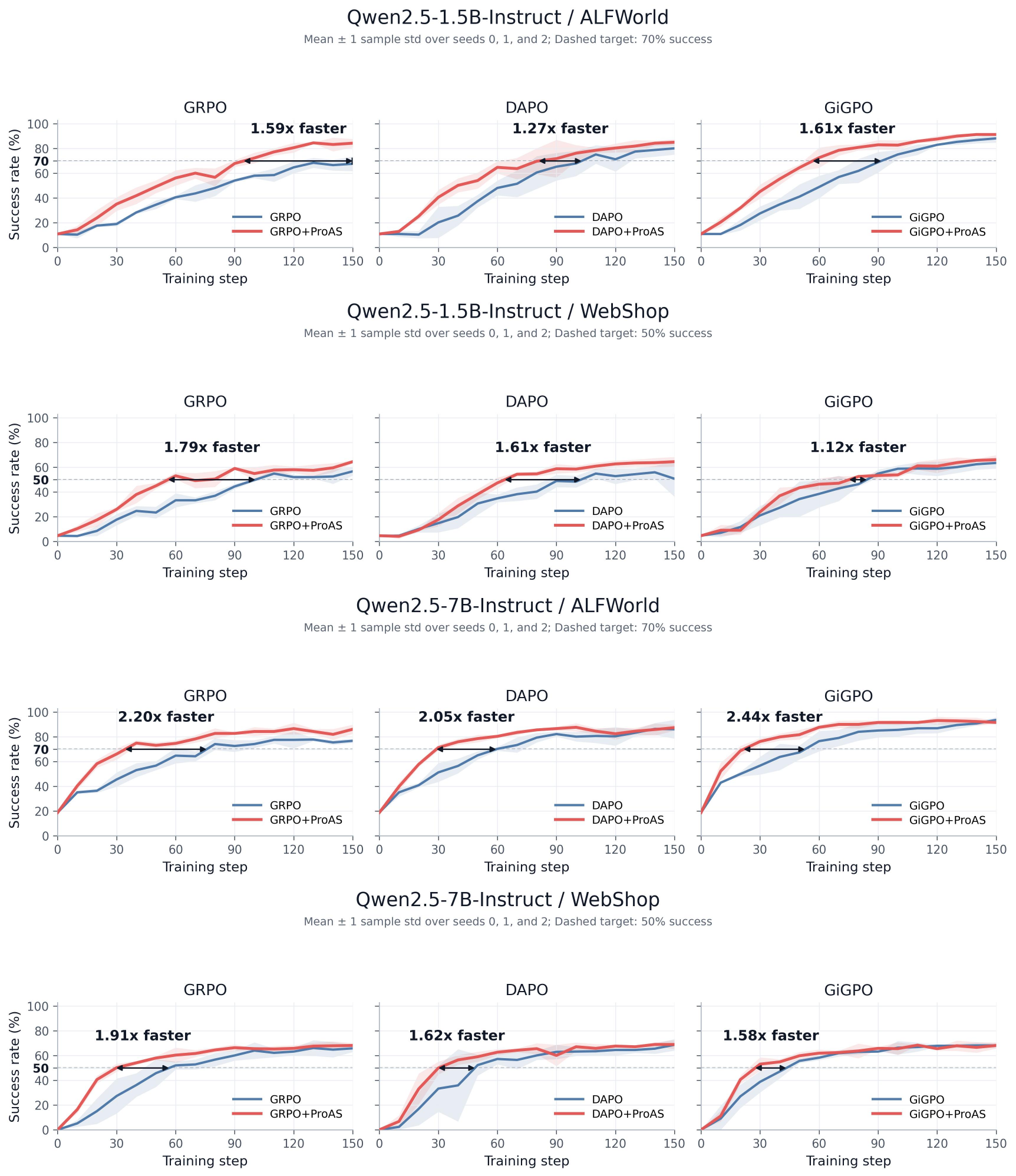}
    \caption{Validation success rate curves on ALFWorld and WebShop within 
     150 RL steps. Dashed horizontal lines mark the fixed success rate
    targets used for computing training speedup: 70 points for ALFWorld and 50
    points for WebShop. Black double arrows indicate the speedup of each base
    algorithm after adding ProAS.}
    \label{fig:training_speedup_overview}
\end{figure*}

Fig.~\ref{fig:training_speedup_overview} visualizes the validation learning
curves used to compute the training-speedup results in Table~\ref{tab:alfworld_webshop}.
For each base algorithm, speedup is measured by the number of RL steps required
to reach a fixed success-rate target before and after adding ProAS.

\clearpage
\newpage
\begin{table*}[t]
    \centering
    \caption{An ALFWorld case illustrating the effect of unreasonable actions
    in the context. The clean and noise-injected conditions
    reach the same continuation state. Each condition contains 50 independently
    sampled continuations. Only the injected unreasonable actions are
    highlighted in pink.}
    \label{tab:alfworld-unreasonable-probing}
    \small
    \renewcommand{\arraystretch}{1.12}
    \setlength{\tabcolsep}{3.5pt}
    \begin{tabularx}{\textwidth}{
        @{}p{1.65cm}
        >{\centering\arraybackslash}p{0.55cm}
        >{\raggedright\arraybackslash}p{3.25cm}
        >{\raggedright\arraybackslash}X
        >{\centering\arraybackslash}p{0.65cm}
        >{\centering\arraybackslash}p{0.55cm}
        @{}
    }
        \toprule
        \multicolumn{6}{@{}l}{\textbf{Case setup}} \\
        \midrule
        Task
        & \multicolumn{5}{>{\raggedright\arraybackslash}p{0.80\textwidth}@{}}{
          Put a keychain on a desk.} \\
        Labels
        & \multicolumn{5}{>{\raggedright\arraybackslash}p{0.80\textwidth}@{}}{
          $\text{Rea}_t=1$ denotes an ARG-supported transition, and
          $\text{PA}_t=1$ denotes advancement to a previously unvisited ARG
          state.} \\
        \midrule
        \multicolumn{6}{@{}l}{\textbf{Task ARG}} \\
        \midrule
        ARG
        & \multicolumn{5}{>{\raggedright\arraybackslash}p{0.80\textwidth}@{}}{
          $\textbf{g}_1$ sidetable reached
          $\xrightarrow{\ \texttt{take keychain 2}\ }$
          $\textbf{g}_2$ keychain held
          $\xrightarrow{\ \texttt{go to desk 1}\ }$
          $\textbf{g}_3$ desk reached while holding the keychain
          $\xrightarrow{\ \texttt{move keychain 2 to desk 1}\ }$
          $\textbf{g}_4$ success.} \\
        Initial edge
        & \multicolumn{5}{>{\raggedright\arraybackslash}p{0.80\textwidth}@{}}{
          The initial transition \texttt{go to sidetable 1} enters
          $\textbf{g}_1$.} \\
        \midrule
        \multicolumn{6}{@{}l}{\textbf{Clean condition}} \\
        \midrule
        Condition
        & $t$
        & Action or operation
        & ARG annotation
        & $\text{Rea}_t$
        & $\text{PA}_t$ \\
        \midrule
        \textbf{Clean}
        & 1
        & \texttt{go to sidetable 1}
        & Enters $\textbf{g}_1$.
        & 1 & 1 \\
        & 2
        & \texttt{take keychain 2 from sidetable 1}
        & Acquires the target keychain and enters $\textbf{g}_2$.
        & 1 & 1 \\
        & 3
        & \texttt{go to desk 1}
        & Reaches the target desk while holding the keychain:
          $\textbf{g}_3$.
        & 1 & 1 \\
        & 4
        & \textit{Continuation sampling} ($n=50$)
        & 50 independently sampled continuations from $\textbf{g}_3$; 41 reach
          $\textbf{g}_4$.
        & -- & -- \\
        & \multicolumn{5}{l}{
          \textit{Expected Success Rate: $82\%$ (41/50);
          Wilson 95\% CI $[69.2\%,90.2\%]$.}} \\
        \midrule
        \multicolumn{6}{@{}l}{\textbf{Noise-injected condition}} \\
        \midrule
        Condition
        & $t$
        & Action or operation
        & ARG annotation
        & $\text{Rea}_t$
        & $\text{PA}_t$ \\
        \midrule
        \textbf{Noise-injected}
        & I1
        & \textcolor{pink!65!black}{\texttt{go to desk 1}}
        & Noise.
        & 0 & 0 \\
        & I2
        & \textcolor{pink!65!black}{\texttt{go to cabinet 1}}
        & Noise.
        & 0 & 0 \\
        & I3
        & \textcolor{pink!65!black}{\texttt{open cabinet 1}}
        & Noise.
        & 0 & 0 \\
        & I4
        & \textcolor{pink!65!black}{\texttt{go to desk 1}}
        & Noise.
        & 0 & 0 \\
        & 1
        & \texttt{go to sidetable 1}
        & Enters $\textbf{g}_1$.
        & 1 & 1 \\
        & 2
        & \texttt{take keychain 2 from sidetable 1}
        & Acquires the target keychain and enters $\textbf{g}_2$.
        & 1 & 1 \\
        & 3
        & \texttt{go to desk 1}
        & Reaches the same continuation state $\textbf{g}_3$.
        & 1 & 1 \\
        & 4
        & \textit{Continuation sampling} ($n=50$)
        & 50 independently sampled continuations from the same $\textbf{g}_3$
          state; 29 reach $\textbf{g}_4$.
        & -- & -- \\
        & \multicolumn{5}{l}{
          \textit{Expected Success Rate: $58\%$ (29/50);
          Wilson 95\% CI $[44.2\%,70.6\%]$.}} \\
        \bottomrule
    \end{tabularx}
\end{table*}

\begin{table*}[t]
    \centering
    \caption{An ALFWorld case comparing exact-action suppression at four
    positions in the same failed rollout. Action indices are 1-based and
    Future-Drift (FD) is the inclusive suffix count of non-advancing actions.
    Repeated non-advancing spans are compressed. Each intervention preserves
    the original prefix, suppresses only the exact original action at the
    selected position, and independently samples 50 continuations.}
    \label{tab:alfworld-varying-fd-suppression}
    \scriptsize
    \renewcommand{\arraystretch}{1.05}
    \setlength{\tabcolsep}{3pt}
    \begin{tabularx}{\textwidth}{
        @{}p{1.90cm}
        >{\centering\arraybackslash}p{0.62cm}
        >{\raggedright\arraybackslash}p{3.25cm}
        >{\raggedright\arraybackslash}X
        >{\centering\arraybackslash}p{0.65cm}
        >{\centering\arraybackslash}p{0.55cm}
        @{}
    }
        \toprule
        \multicolumn{6}{@{}l}{\textbf{Case setup}} \\
        \midrule
        Task
        & \multicolumn{5}{>{\raggedright\arraybackslash}p{0.78\textwidth}@{}}{
          Look at a bowl under the desk lamp.} \\
        Labels
        & \multicolumn{5}{>{\raggedright\arraybackslash}p{0.78\textwidth}@{}}{
          $\text{Rea}_t=1$ denotes an ARG-supported transition, and
          $\text{PA}_t=1$ denotes advancement to a previously unvisited ARG
          state. FD counts the current non-advancing action and all subsequent
          non-advancing actions in the original trajectory.} \\
        Intervention
        & \multicolumn{5}{>{\raggedright\arraybackslash}p{0.78\textwidth}@{}}{
          At the selected decision, an exact match to the original action is
          rejected and that decision is resampled; all other candidate actions
          remain eligible.} \\
        \midrule
        \multicolumn{6}{@{}l}{\textbf{Task ARG}} \\
        \midrule
        ARG
        & \multicolumn{5}{>{\raggedright\arraybackslash}p{0.78\textwidth}@{}}{
          Initial state
          $\xrightarrow{\ \texttt{go to desk 1}\ }$
          $\textbf{g}_1$ desk reached
          $\xrightarrow{\ \texttt{take bowl 1/2}\ }$
          $\textbf{g}_2$ target bowl held
          $\xrightarrow{\ \texttt{use desklamp 1}\ }$
          $\textbf{g}_3$ success.} \\
        \midrule
        \multicolumn{6}{@{}l}{\textbf{Raw trajectory}} \\
        \midrule
        Condition
        & $t$
        & Action or operation
        & ARG annotation
        & $\text{Rea}_t$
        & $\text{PA}_t$ \\
        \midrule
        \textbf{Original failure}
        & 1--2
        & \texttt{look} $\times 2$
        & Initial probe followed by a self-loop; no ARG progress.
        & 0 & 0 \\
        & 3
        & \texttt{look} \quad (FD $=47$)
        & Admissible self-loop at the first intervention position.
        & 0 & 0 \\
        & 4--7
        & \texttt{look} $\times 4$
        & Repeated self-loops before reaching the desk.
        & 0 & 0 \\
        & 8
        & \texttt{go to desk 1}
        & Enters $\textbf{g}_1$; the trajectory's only advancing action.
        & 1 & 1 \\
        & 9--27
        & \texttt{look} $\times 15$; probes; one unsupported bowl command
        & Nineteen non-advancing actions remain at $\textbf{g}_1$.
        & 0 & 0 \\
        & 28
        & \texttt{look at bowl under desklamp 1} \quad (FD $=23$)
        & Inadmissible command at the second intervention position.
        & 0 & 0 \\
        & 29--39
        & Repeated bowl/lamp commands, probes, and redundant navigation
        & Eleven non-advancing actions remain at $\textbf{g}_1$.
        & 0 & 0 \\
        & 40
        & \texttt{look at bowl 1 from desk 1 with desklamp 1}
          \quad (FD $=11$)
        & Inadmissible command at the third intervention position.
        & 0 & 0 \\
        & 41--46
        & Malformed or unsupported bowl/lamp commands and probes
        & Six non-advancing actions remain at $\textbf{g}_1$.
        & 0 & 0 \\
        & 47
        & \texttt{look at bowl 2} \quad (FD $=4$)
        & Inadmissible command at the final intervention position.
        & 0 & 0 \\
        & 48--50
        & \texttt{look at bowl 2} $\times 3$
        & Repeated inadmissible actions exhaust the horizon.
        & 0 & 0 \\
        & \multicolumn{5}{l}{\textit{Outcome: failure after the 50-step horizon.}} \\
        \midrule
        \multicolumn{6}{@{}l}{\textbf{Suppressing on Step 3 (Future-Drift 47)}} \\
        \midrule
        \textbf{Step 3}
        & 1--2
        & Replay the original prefix
        & Returns to the decision state immediately before Step 3.
        & -- & -- \\
        & 3
        & Suppress exact action \texttt{look}
        & Reject and resample only this action at the selected decision.
        & 0 & 0 \\
        & 3--50
        & \textit{Continuation sampling} ($n=50$)
        & 12 independently sampled continuations succeed.
        & -- & -- \\
        & \multicolumn{5}{l}{
          \textit{Expected Success Rate: $24\%$ (12/50); Wilson 95\% CI
          $[14.3\%,37.4\%]$.}} \\
        \midrule
        \multicolumn{6}{@{}l}{\textbf{Suppressing on Step 28 (Future-Drift 23)}} \\
        \midrule
        \textbf{Step 28}
        & 1--27
        & Replay the original prefix
        & Returns to the decision state immediately before Step 28.
        & -- & -- \\
        & 28
        & Suppress exact original bowl/lamp command
        & Reject and resample only the selected inadmissible action.
        & 0 & 0 \\
        & 28--50
        & \textit{Continuation sampling} ($n=50$)
        & 8 independently sampled continuations succeed.
        & -- & -- \\
        & \multicolumn{5}{l}{
          \textit{Expected Success Rate: $16\%$ (8/50); Wilson 95\% CI
          $[8.3\%,28.5\%]$.}} \\
        \midrule
        \multicolumn{6}{@{}l}{\textbf{Suppressing on Step 40 (Future-Drift 11)}} \\
        \midrule
        \textbf{Step 40}
        & 1--39
        & Replay the original prefix
        & Returns to the decision state immediately before Step 40.
        & -- & -- \\
        & 40
        & Suppress exact original bowl/lamp command
        & Reject and resample only the selected inadmissible action.
        & 0 & 0 \\
        & 40--50
        & \textit{Continuation sampling} ($n=50$)
        & 5 independently sampled continuations succeed.
        & -- & -- \\
        & \multicolumn{5}{l}{
          \textit{Expected Success Rate: $10\%$ (5/50); Wilson 95\% CI
          $[4.3\%,21.4\%]$.}} \\
        \midrule
        \multicolumn{6}{@{}l}{\textbf{Suppressing on Step 47 (Future-Drift 4)}} \\
        \midrule
        \textbf{Step 47}
        & 1--46
        & Replay the original prefix
        & Returns to the decision state immediately before Step 47.
        & -- & -- \\
        & 47
        & Suppress exact action \texttt{look at bowl 2}
        & Reject and resample only the selected inadmissible action.
        & 0 & 0 \\
        & 47--50
        & \textit{Continuation sampling} ($n=50$)
        & None of the 50 independently sampled continuations succeeds.
        & -- & -- \\
        & \multicolumn{5}{l}{
          \textit{Expected Success Rate: $0\%$ (0/50); Wilson 95\% CI
          $[0.0\%,7.1\%]$.}} \\
        \bottomrule
    \end{tabularx}
\end{table*}

\begin{table*}[!t]
  \centering
  \caption{An ALFWorld case illustrating strict ARG purification and online
  graph evolution. The action-level annotations are computed against the
  pre-update graph $\mathcal{G}^{(17)}$. After the rollout succeeds, only its
  valid, non-self-loop transitions are extracted as a new reference path and
  merged into $\mathcal{G}^{(18)}$.}
  \label{tab:alfworld-arg-purity}
  \small
  \renewcommand{\arraystretch}{1.12}
  \setlength{\tabcolsep}{3.5pt}
  \begin{tabularx}{\textwidth}{
      @{}>{\centering\arraybackslash}p{1.10cm}
      >{\raggedright\arraybackslash}p{2.85cm}
      >{\raggedright\arraybackslash}X
      >{\centering\arraybackslash}p{0.62cm}
      >{\centering\arraybackslash}p{0.55cm}
      >{\raggedright\arraybackslash}p{1.55cm}@{}
  }
      \toprule
      \multicolumn{6}{@{}l}{\textbf{Case setup}} \\
      \midrule
      Task
      & \multicolumn{5}{>{\raggedright\arraybackslash}p{0.87\textwidth}@{}}{
        Look at a pencil under the desk lamp.} \\
      Source
      & \multicolumn{5}{>{\raggedright\arraybackslash}p{0.87\textwidth}@{}}{
        Qwen2.5-1.5B-Instruct rollout at training iteration 17; success; reward $=10$.} \\
      $\mathcal{G}^{(17)}$
      & \multicolumn{5}{>{\raggedright\arraybackslash}p{0.87\textwidth}@{}}{
        Existing expert path:
        initial state
        $\xrightarrow{\ \texttt{go to desk 1}\ }$
        $\textbf{g}_1$ desk reached
        $\xrightarrow{\ \texttt{use desklamp 1}\ }$
        $\textbf{g}_2$ desk lamp activated
        $\xrightarrow{\ \texttt{take pencil 2 from desk 1}\ }$
        $\textbf{g}_3$ success.} \\
      Labels
      & \multicolumn{5}{>{\raggedright\arraybackslash}p{0.87\textwidth}@{}}{
        $\text{Rea}_t=1$ denotes a transition supported by
        $\mathcal{G}^{(17)}$, and $\text{PA}_t=1$ denotes advancement to a
        previously unvisited ARG state. The update decision is applied only
        after environment-verified success.} \\
      \midrule
      \multicolumn{6}{@{}l}{\textbf{Raw successful rollout and ARG annotations}} \\
      \midrule
      $t$ & Action $a_t$ & Annotation under $\mathcal{G}^{(17)}$
      & $\text{Rea}_t$ & $\text{PA}_t$ & Purification \\
      \midrule
      1 & \texttt{go to desk 1}
      & Matches the expert transition and reaches $\textbf{g}_1$.
      & 1 & 1 & Retain \\
      2 & \texttt{examine desk 1}
      & {Invalid action}.
      & 0 & 0 & Prune \\
      3 & \texttt{examine desk 1}
      & Repeats the same abstract state (\texttt{self\_loop}).
      & 0 & 0 & Prune \\
      4 & \texttt{examine desk 1}
      & Repeats the same abstract state (\texttt{self\_loop}).
      & 0 & 0 & Prune \\
      5 & \texttt{examine desk 1}
      & Repeats the same abstract state (\texttt{self\_loop}).
      & 0 & 0 & Prune \\
      6 & \texttt{look at \#pencil 1 with \#desklamp 1}
      & Inadmissible action.
      & 0 & 0 & Prune \\
      7 & \texttt{inventory}
      & Redundant examination.
      & 0 & 0 & Prune \\
      8 & \texttt{use desklamp 1}
      & Matches the expert transition and reaches $\textbf{g}_2$.
      & 1 & 1 & Retain \\
      9 & \texttt{examine desk 1}
      & Redundant examination.
      & 0 & 0 & Prune \\
      10 & \texttt{take pencil 1 from desk 1}
      & Novel and valid action toward success.
      & 0 & 0 & Retain  \\
        \midrule
        \multicolumn{6}{@{}l}{\textbf{Evolved ARG}} \\
        \midrule
        \multicolumn{6}{@{}c@{}}{
          \begin{tikzpicture}[x=1cm,y=1cm,>=stealth,
              state/.style={font=\scriptsize,align=center},
              edge label/.style={font=\tiny\ttfamily,fill=white,
                inner sep=1.5pt}]
            \node[state,anchor=east] at (-0.45,0)
              {$\mathcal{G}^{(18)}$};
            \node[state] (init) at (0.45,0) {initial\\state};
            \node[state] (desk) at (2.85,0)
              {$\textbf{g}_1$\\desk reached};
            \node[state] (lamp) at (5.75,0)
              {$\textbf{g}_2$\\desk lamp activated};
            \node[state] (expert) at (10.75,0)
              {$\textbf{g}_3$\\\texttt{pencil 2} acquired\\
               \textit{(existing expert path)}};
            \node[state] (online) at (10.75,-1.18)
              {$\textbf{g}'_3$\\\texttt{pencil 1} acquired\\
               \textit{(new online path)}};

            \draw[->] (init.east) -- node[edge label,above]
              {go to desk 1} (desk.west);
            \draw[->] (desk.east) -- node[edge label,above]
              {use desklamp 1} (lamp.west);
            \draw[->] (lamp.east) --
              node[edge label,above,pos=0.40]
              {take pencil 2 from desk 1} (expert.west);
            \draw[->] (lamp.south) -- (5.75,-1.18) --
              node[edge label,below]
              {take pencil 1 from desk 1} (online.west);
          \end{tikzpicture}
        } \\
        \bottomrule
  \end{tabularx}
\end{table*}


\begin{table*}[t]
    \centering
    \caption{A WebShop case illustrating order-invariant option completion and
    action-level ProAS annotations. Trajectories A and B complete the option
    sub-goal in different orders and both succeed; Trajectory C shows
    a failed rollout with unreasonable or non-advancing actions.}
    \label{tab:webshop-option-permutation}
    \small
    \renewcommand{\arraystretch}{1.12}
    \setlength{\tabcolsep}{3.5pt}
    \begin{tabularx}{\textwidth}{
        @{}p{1.65cm}
        >{\centering\arraybackslash}p{0.55cm}
        >{\raggedright\arraybackslash}p{3.05cm}
        >{\raggedright\arraybackslash}X
        >{\centering\arraybackslash}p{0.65cm}
        >{\centering\arraybackslash}p{0.55cm}
        @{}
    }
        \toprule
        \multicolumn{6}{@{}l}{\textbf{Case setup}} \\
        \midrule
        Task
        & \multicolumn{5}{>{\raggedright\arraybackslash}p{0.80\textwidth}@{}}{
          Find women's activewear with nylon spandex, \texttt{color=green},
          \texttt{size=x-large}, and price below \$30.} \\
        Target
        & \multicolumn{5}{>{\raggedright\arraybackslash}p{0.80\textwidth}@{}}{
          \texttt{B08G8DXR5N}} \\
        ARG
        & \multicolumn{5}{>{\raggedright\arraybackslash}p{0.80\textwidth}@{}}{
          $\textbf{g}_1$ target visible
          $\rightarrow$ $\textbf{g}_2$ target opened
          $\rightarrow$ $\textbf{g}_3$ options \{\texttt{green}, \texttt{x-large}\}
          $\rightarrow$ $\textbf{g}_4$ price verified
          $\rightarrow$ $\textbf{g}_5$ attributes verified
          $\rightarrow$ $\textbf{g}_6$ purchase.} \\
        Labels
        & \multicolumn{5}{>{\raggedright\arraybackslash}p{0.80\textwidth}@{}}{
          $\text{Rea}_t=1$ denotes an ARG-supported transition, and
          $\text{PA}_t=1$ denotes entry into a previously unvisited valid ARG
          state. Thus, $(1,1)$ is productive, $(1,0)$ is reasonable but
          non-advancing, and $(0,0)$ is unsupported.} \\
        \midrule
        \multicolumn{6}{@{}l}{\textbf{Rollout comparison}} \\
        \midrule
        Trajectory
        & $t$
        & Action $a_t$
        & ARG annotation
        & $\text{Rea}_t$
        & $\text{PA}_t$ \\
        \midrule
        \textbf{A: size first}
        & 1
        & \texttt{search[\ldots]}
        & Target product appears in the search results: $s_{1,1}\in\textbf{g}_1$.
        & 1 & 1 \\
        & 2
        & \texttt{click[target]}
        & Target product page is opened: $s_{2,1}\in\textbf{g}_2$.
        & 1 & 1 \\
        & 3
        & \texttt{click[x-large]}
        & Partial $\textbf{g}_3$ state: the required size is selected.
        & 1 & 1 \\
        & 4
        & \texttt{click[green]}
        & $\textbf{g}_3$ completed: all required options are selected.
        & 1 & 1 \\
        & 5
        & \texttt{click[buy now]}
        & Price and attributes are verified; purchase reaches $\textbf{g}_6$.
        & 1 & 1 \\
        & \multicolumn{5}{l}{\textit{Outcome: success; reward $=10$.}} \\
        \midrule
        \textbf{B: color first}
        & 1
        & \texttt{search[\ldots]}
        & Target product appears in the search results: $s_{1,1}\in\textbf{g}_1$.
        & 1 & 1 \\
        & 2
        & \texttt{click[target]}
        & Target product page is opened: $s_{2,1}\in\textbf{g}_2$.
        & 1 & 1 \\
        & 3
        & \texttt{click[green]}
        & Partial $\textbf{g}_3$ state: the required color is selected.
        & 1 & 1 \\
        & 4
        & \texttt{click[x-large]}
        & $\textbf{g}_3$ completed: all required options are selected.
        & 1 & 1 \\
        & 5
        & \texttt{click[buy now]}
        & Price and attributes are verified; purchase reaches $\textbf{g}_6$.
        & 1 & 1 \\
        & \multicolumn{5}{l}{\textit{Outcome: success; reward $=10$.}} \\
        \midrule
        \textbf{C: failed}
        & 1
        & \texttt{search[\ldots]}
        & Target product appears in the search results: $s_{1,1}\in\textbf{g}_1$.
        & 1 & 1 \\
        & 2
        & \texttt{click[target]}
        & Target product page is opened: $s_{2,1}\in\textbf{g}_2$.
        & 1 & 1 \\
        & 3
        & \texttt{click[large]}
        & Wrong size option; the transition is not supported by the ARG.
        & 0 & 0 \\
        & 4
        & \texttt{click[description]}
        & Neutral page inspection that does not satisfy any ARG requirement.
        & 0 & 0 \\
        & 5
        & \texttt{click[buy now]}
        & Premature purchase before the required option set is completed.
        & 0 & 0 \\
        & 9
        & \texttt{click[x-large]}
        & Recovery segment reaches a partial $\textbf{g}_3$ state.
        & 1 & 1 \\
        & 10
        & \texttt{click[green]}
        & $\textbf{g}_3$ completed after the required color is selected.
        & 1 & 1 \\
        & 11
        & \texttt{click[x-large]}
        & Redundant option click; the valid ARG state has already been visited.
        & 1 & 0 \\
        & \multicolumn{5}{l}{\textit{Outcome: failure; reward $=0$.}} \\
        \bottomrule
    \end{tabularx}
\end{table*}


\clearpage
\newpage
\begin{table}[H]
\centering
\caption{A turn-level SFT example from ALFWorld. Underlined text denotes
instance-specific fields populated from the task, environment state, and
interaction history.}
\label{tab:sft_data_alfworld}
\begin{minipage}{0.96\linewidth}
\begin{promptbox}{SFT Example for ALFWorld}
\footnotesize\raggedright

\textcolor{blue!80!black}{\textbf{System Prompt:}} \\
You are Qwen, created by Alibaba Cloud. You are a helpful assistant.

\vspace{1mm}
{\color{gray!40}\hrule}
\vspace{1mm}

\textcolor{blue!80!black}{\textbf{User Prompt:}} \\
You are an expert agent operating in the ALFRED Embodied Environment. Your task
is to: \filled{put two book in bed}.\\
Prior to this step, you have already taken \filled{4} step(s). Below are the most recent
\filled{2} observations and the corresponding actions you took:\\
\filled{{[}Observation 3: `You arrive at sidetable 1. On the sidetable 1, you see an
alarmclock 1, a cd 2, and a pencil 2.', Action 3: `go to shelf 1'{]}}\\
\filled{{[}Observation 4: `You arrive at shelf 1. On the shelf 1, you see nothing.',
Action 4: `go to drawer 1'{]}}\\
You are now at step \filled{5} and your current observation is: \filled{You arrive at drawer 1.
The drawer 1 is closed}.\\
Your admissible actions of the current situation are: \filled{[`examine drawer 1', `go
to bed 1', `go to desk 1', `go to drawer 2', `go to garbagecan 1', `go to
shelf 1', `go to sidetable 1', `inventory', `look', `open drawer 1']}.

Now it's your turn to take an action.\\
You should first reason step-by-step about the current situation. This reasoning
process MUST be enclosed within \thinktag{think} \thinktag{/think} tags.\\
Once you've finished your reasoning, you should choose an admissible action for
the current step and present it within \actiontag{action} \actiontag{/action}
tags.

\vspace{1mm}
{\color{gray!40}\hrule}
\vspace{1mm}

\textcolor{blue!80!black}{\textbf{Assistant Response (Supervision Target):}} \\
\thinktag{think} The best action is to open drawer 1. \thinktag{/think}\\
\actiontag{action} open drawer 1 \actiontag{/action}
\end{promptbox}
\end{minipage}
\end{table}

\begin{table}[H]
\centering
\caption{A turn-level SFT example from WebShop. Underlined text denotes
instance-specific fields populated from the task, environment state, and
interaction history.}
\label{tab:sft_data_webshop}
\begin{minipage}{0.96\linewidth}
\begin{promptbox}{SFT Example for WebShop}
\footnotesize\raggedright

\textcolor{blue!80!black}{\textbf{System Prompt:}} \\
You are Qwen, created by Alibaba Cloud. You are a helpful assistant.

\vspace{1mm}
{\color{gray!40}\hrule}
\vspace{1mm}

\textcolor{blue!80!black}{\textbf{User Prompt:}} \\
You are an expert autonomous agent operating in the WebShop e-commerce
environment.\\
Your task is to: \filled{Find me plug play, easy use turntables, and price lower than
150.00 dollars}.\\
Prior to this step, you have already taken \filled{1} step. Below is the most recent
observation and the corresponding action you took:\\
\filled{{[}Observation 1: `WebShop [SEP] Instruction: [SEP] Find me plug play, easy use
turntables, and price lower than 150.00 dollars [SEP] Search', Action 1:
`search[turntables]'{]}}\\
You are now at step \filled{2} and your current observation is: \filled{`Back to Search' [SEP]
`Page 1 (Total results: 1)' [SEP] `Next $>$' [SEP] `B09Q6CZ59K' [SEP]
`ngruama8 USB Portable Mini Vinyl Turntable Audio Player' [SEP] `\$110.54'}.\\
Your admissible actions of the current situation are: \filled{[`click[back to search]',
`click[next $>$]', `click[b09q6cz59k]']}.

Now it's your turn to take one action for the current step.\\
You should first reason step-by-step about the current situation, then determine
which admissible action best advances the shopping goal. This reasoning process
MUST be enclosed within \thinktag{think} \thinktag{/think} tags.\\
Once you've finished your reasoning, you should choose an admissible action for
the current step and present it within \actiontag{action} \actiontag{/action}
tags.

\vspace{1mm}
{\color{gray!40}\hrule}
\vspace{1mm}

\textcolor{blue!80!black}{\textbf{Assistant Response (Supervision Target):}} \\
\thinktag{think} Open the target product B09Q6CZ59K. \thinktag{/think}\\
\actiontag{action} click[b09q6cz59k] \actiontag{/action}
\end{promptbox}
\end{minipage}
\end{table}

\end{document}